\documentclass[sigconf]{acmart}

\usepackage[utf8]{inputenc} 
\usepackage[T1]{fontenc}    
\usepackage{hyperref}       
\usepackage{url}            
\usepackage{booktabs}       
\usepackage{amsfonts}       
\usepackage{nicefrac}       
\usepackage{microtype}      
\usepackage{xcolor}         
\usepackage{graphicx}
\usepackage{amsmath}
\usepackage{amsthm}
\usepackage{booktabs}
\usepackage{algorithm}
\usepackage{algpseudocode}
\usepackage{multirow}
\usepackage{subfigure}
\usepackage{listings}
\usepackage{threeparttable}
\usepackage{wrapfig}
\usepackage{diagbox}
\usepackage{pifont}

\newtheorem{definition}{Definition}
\newtheorem{proposition}{Proposition}
\newtheorem{assumption}{Assumption}
\newtheorem{remark}{Remark}

\definecolor{gray}{rgb}{0.9, 0.9, 0.9}

\definecolor{pink}{rgb}{0.858, 0.188, 0.478}
\definecolor{commentcolor}{RGB}{110,154,155}   

\AtBeginDocument{%
  \providecommand\BibTeX{{%
    \normalfont B\kern-0.5em{\scshape i\kern-0.25em b}\kern-0.8em\TeX}}}

\setcopyright{acmcopyright}
\copyrightyear{2023}
\acmYear{2023}
\setcopyright{acmlicensed}\acmConference[CIKM '23]{Proceedings of the 32nd ACM International Conference on Information and Knowledge Management}{October 21--25, 2023}{Birmingham, United Kingdom}
\acmBooktitle{Proceedings of the 32nd ACM International Conference on Information and Knowledge Management (CIKM '23), October 21--25, 2023, Birmingham, United Kingdom}
\acmPrice{15.00}
\acmDOI{10.1145/3583780.3614903}
\acmISBN{979-8-4007-0124-5/23/10}
\begin{document}

\title{\textsc{Guard}: Graph Universal Adversarial Defense}


\author{Jintang Li}
\affiliation{\institution{Sun Yat-sen University}
    \country{}}
\email{lijt55@mail2.sysu.edu.cn}

\author{Jie Liao}
\affiliation{\institution{Sun Yat-sen University}
    \country{}}
\email{liaoj27@mail2.sysu.edu.cn}

\author{Ruofan Wu}
\affiliation{\institution{Ant Group}
    \country{}}
\email{ruofan.wrf@antgroup.com}

\author{Liang Chen}
\authornote{Corresponding author.}
\affiliation{\institution{Sun Yat-sen University}
    \country{}}
\email{chenliang6@mail.sysu.edu.cn}

\author{Zibin Zheng}
\affiliation{\institution{Sun Yat-sen University}
    \country{}}
\email{zhzibin@mail.sysu.edu.cn}

\author{Jiawang Dan}
\affiliation{\institution{Ant Group}
    \country{}}
\email{yancong.djw@antgroup.com}

\author{Changhua Meng}
\affiliation{\institution{Ant Group}
    \country{}}
\email{changhua.mch@antgroup.com}

\author{Weiqiang Wang}
\affiliation{\institution{Ant Group}
    \country{}}
\email{weiqiang.wwq@antgroup.com}

\renewcommand{\shortauthors}{Jintang Li et al.}


\begin{abstract}
    Graph convolutional networks (GCNs) have been shown to be vulnerable to small adversarial perturbations, which becomes a severe threat and largely limits their applications in security-critical scenarios. To mitigate such a threat, considerable research efforts have been devoted to increasing the robustness of GCNs against adversarial attacks. However, current defense approaches are typically designed to prevent GCNs from \textit{untargeted} adversarial attacks and focus on overall performance, making it challenging to protect important local nodes from more powerful \textit{targeted} adversarial attacks. Additionally, a trade-off between robustness and performance is often made in existing research.
    Such limitations highlight the need for developing an effective and efficient approach that can defend local nodes against targeted attacks, without compromising the overall performance of GCNs.
    In this work, we present a simple yet effective method, named \textbf{\underline{G}}raph \textbf{\underline{U}}niversal \textbf{\underline{A}}dve\textbf{\underline{R}}sarial \textbf{\underline{D}}efense (\textsc{Guard}). Unlike previous works, \textsc{Guard} protects each individual node from attacks with a \textit{universal} defensive patch, which is generated once and can be applied to any node (\textit{node-agnostic}) in a graph.
    \textsc{Guard} is fast, straightforward to implement without any change to network architecture nor any additional parameters, and is broadly applicable to any GCNs.
    Extensive experiments on four benchmark datasets demonstrate that \textsc{Guard} significantly improves robustness for several established GCNs against multiple adversarial attacks and outperforms state-of-the-art defense methods by large margins.
\end{abstract}



\maketitle

\section{Introduction}
Graph structured data are ubiquitous in real world, with prominent examples including financial networks~\cite{DBLP:conf/icdm/WangQL0JWFYZY19}, molecular fingerprints~\cite{DBLP:conf/ijcai/ZhaoLHLZ21}, and recommender systems~\cite{DBLP:conf/sigir/WuWF0CLX21}. Graph convolutional networks (GCNs)~\cite{DBLP:conf/iclr/KipfW17}, a series of neural network models primarily developed for graph structured data, have met with great success in a variety of applications and domains. This is mainly due to their great capacity in jointly leveraging information from both graph structure and node features. Over the past few years, research on GCNs has surged to become one of the hottest topics in deep learning community~\cite{dlgsurvey_tkde20}.

Despite the success of GCNs in numerous graph-based machine learning tasks, \textit{e.g.}, link prediction and node classification, they suffer seriously from vulnerability to adversarial attacks. As shown in~\cite{DBLP:conf/kdd/ZugnerAG18,li2021adversarial}, slight perturbations on either node features or graph structure can lead to incorrect predictions of GCNs on specific nodes. This attack is also known as \textit{targeted attack}~\cite{DBLP:conf/kdd/ZugnerAG18}. Even worse, \cite{gua} has recently shown the possibility of misleading GCNs' classification on ``any'' target node by performing a \textit{node-agnostic, universal} adversarial perturbation.

The adversarial targeted attack is a real threat. For example, it provides a possibility of enabling a fraudster to disguise himself as a regular user to bypass GCNs based anti-fraud systems and disperse disinformation or reap end-users' privacy~\cite{DBLP:conf/cikm/DouL0DPY20}. Hence, the need for countermeasures against such an attack becomes more critical. So far, heuristics have been investigated in the literature to mitigate the risk of adversarial attacks from different ways~\cite{chen2020survey}. Among contemporary approaches, one of the most simple and effective ways of defense is to preprocess the graph and alleviate the adversarial behaviors in advance. In this regard, \cite{Wu0TDLZ19} first propose to remove suspicious edges between suspicious nodes based on Jaccard similarity, \cite{EntezariADP20} leverage SVD to form a low-rank approximation of graph to reduce the effect of attacks. Current works can indeed protect GCNs from attacks toward the whole network. Their defenses, however, are typically designed for the whole graph while ignoring the protection of important local nodes, making them suffer seriously from stronger adversarial targeted attacks.
In addition, there is often a particular tradeoff between performance and robustness since they often hold the assumption that data has already been poisoned.

In this work, we consider a more practical and flexible defense strategy meant to generate what we term as \textit{universal} defensive patch that can be applied to an arbitrary node. The defensive patch can be performed on the graph as an active defense, which mitigates the risk of adversarial attacks at test time by removing malicious edges from input graphs. Here a defensive patch is a \texttt{0}-\texttt{1} binary vector, where \texttt{1} denotes the edge modification (\textit{e.g.,} removal) and \texttt{0} otherwise. To our best knowledge, we are the first to study universal defense on graphs.

\textbf{Why universal defense?}
As a new perspective of defense strategy, our work is orthogonal to recent studies on robustifying GCNs. The universal defense is advantageous as (i) no access to the target node or victim model is needed at test time, and (ii) it drastically lowers the barrier to defend against adversarial attacks: the universal defensive patch is generated once and can be applied to any model (\textit{model-agnostic}) and any node (\textit{node-agnostic}) in a graph.

\textbf{Is universal defense achievable?}
Although there have been several successful attempts in vision research, the possibility of a universal defense against adversarial targeted attacks on graphs remains largely unexplored. Particularly, graphs often come with complicated structures, in which relationships between nodes can be challenging to capture. As a result, achieving a universal defense may not be a straightforward task.
In this paper, we seek to uncover the intrinsic patterns of adversarial attacks, and one step further, explore the feasibility of universal defenses on graphs.
Specifically, we empirically discover an interesting phenomenon that attackers prefer picking the same attacker nodes from a set of low-degree nodes when perturbing different target nodes. The finding suggests that these nodes may be part of what makes the model vulnerable, and thus a universal defense becomes possible if one can identify them in advance.

In light of above insights, the core problem we raise and address in this paper is:
\begin{center}
    \textit{How to design a universal defense that works for any individual node to defend against adversarial attacks?}
\end{center}

In this paper, we propose \textbf{\underline{G}}raph \textbf{\underline{U}}niversal \textbf{\underline{A}}dve\textbf{\underline{R}}sarial \textbf{\underline{D}}efense (\textsc{Guard}) to address this problem for the first time. Unlike previous works, \textsc{Guard} applies a universal patch to protect any node from adversarial targeted attacks without knowledge of victim GCNs. We empirically show that our method can significantly improve the robustness of victim GCNs against a variety of adversarial targeted attacks. Particularly, we demonstrate that GCNs equipped with \textsc{Guard} can also outperform current state-of-the-art defenses with large margins.

This paper offers the following main contributions:
\begin{itemize}
    \item We demonstrate, both theoretically and empirically, that current attacks tend to perturb a target node with a fixed set of low-degree nodes. Our finding offers deeper insights on understanding the vulnerability of GCNs.
    \item We propose \textsc{Guard}, an effective and scalable universal defense to protect an arbitrary node from multiple adversarial attacks. \textsc{Guard} comes with good generality and flexibility for well-established GCNs, enabling them to be robust against adversarial attacks.
    \item Extensive experiments on four public graph benchmarks demonstrate that \textsc{Guard} can protect GCNs from strong adversarial targeted attacks without sacrificing the clean performance.
\end{itemize}

To our best knowledge, \textsc{Guard} is the first successful attempt in applying universal defense on graphs.
We believe that our work is a step forward in the development of simple and provably effective defenses, and hope that it will inspire both theoretical and practical future research efforts.


\section{Related Work}\label{sec:related_work}
The robustness of graph convolutional networks against adversarial attacks has gained increasing attention in the last few years~\cite{DBLP:conf/kdd/ZugnerAG18,DBLP:conf/iclr/ZugnerG19,DBLP:conf/kdd/ZhuZ0019,li2021adversarial,chen2021understanding,reliable_dgl}. While there are numerous (heuristic) approaches aimed at robustifying GCNs, there is always a newly devised stronger attack attempts to break them, leading to an arms race between attackers and defenders~\cite{chen2020survey}.

\paragraph{Adversarial attack on graphs.}
Literature is rich on attacking GCNs with adversarial examples. The most widely used solution for crafting adversarial examples on graphs is to utilize a locally trained surrogate model (typically a GCN)~\cite{DBLP:conf/kdd/ZugnerAG18}. In this way, an attacker obtains the approximated gradient/loss towards edges or edge modifications in a graph~\cite{DBLP:conf/iclr/ZugnerG19,rbcd,chen2018fast,Wu0TDLZ19} or subgraph~\cite{li2021adversarial} to craft the worst-case perturbations, and subsequently transfer them to other victim models as a practical gray-box attack. These surrogate-based attacks have become a serious threat to GCNs because they are able to attack the target GCNs without requiring sufficient knowledge of them.

\paragraph{Adversarial defense on graphs.}
Extensive research efforts have been made on improving the robustness of GCNs, which can be typically classified into three categories: (i) robust training (\textit{e.g.,} adversarial training)~\cite{XuC0CWHL19,sat}, (ii) model robustification that focuses on either the message passing scheme or the network architecture~\cite{chen2021understanding,DBLP:conf/kdd/ZhuZ0019,simpgcn} and (iii) graph preprocessing~\cite{Wu0TDLZ19,EntezariADP20}. A general drawback shared by previous methods is the lack of scalability, which makes them less capable of dealing with graphs that are substantially larger than PubMed~\cite{DBLP:journals/aim/SenNBGGE08}. In addition, the aforementioned defenses primarily aim to mitigate attacks on the \textit{entire} network, rather than protecting GCNs from \textit{local} targeted adversarial attacks.



\paragraph{Universal attack and defense.}
Recent works show a new trend of attacking neural networks by universal adversarial attacks~\cite{Moosavi-Dezfooli17,ijcai2021-635}, \textit{i.e.}, unique perturbations that transfer across different inputs. The universal attacks have been widely studied in vision research. Until recently, \cite{gua} first extend the idea to graph domain by crafting a single and universal perturbation that is capable to fool a GCN when applied to any target node. Another line of research is universal defense, which devises a universal `watermark' to protect neural networks from multiple attacks~\cite{huang2021cmuawatermark}. Despite the recent interest in problems of universal defense among vision research, there has been relatively little work that explores the universal defense in the graph domain.

\section{Preliminaries}
\subsection{Notations}
In line with the focus of our work, we briefly outline necessary definitions used throughout this paper.
Let $\mathcal{G}=(\mathcal{V},\mathcal{E})$ be an undirected graph, with the node set $\mathcal{V}=\{v_1, \ldots, v_N\}$ and the undirected edge set $\mathcal{E}=\{e_1,\ldots,e_M\}$. The corresponding adjacency matrix is denoted as $A\in \{0,1\}^{N\times N}$ with $(u, v)$ entry equaling to \texttt{1} if there is an edge between $u$ and $v$ and \texttt{0} otherwise. We denote $d$ the node degrees where $d_u=\sum_v A_{u,v}$. Also, each node is associated with an $F$-dimensional feature vector and $X \in  \mathbb{R} ^{N \times F}$ denotes the feature matrix for all $N$ nodes. In the node classification task, each node $v$ is associated with one class label $y_v \in \mathcal{C}$ where $\mathcal{C}$ is the set of all candidate classes. Given a subset of nodes $\mathcal{V}_\text{train} \subset \mathcal{V}$ are labeled, the goal is to learn a function $f_\theta$ that maps each node $v\in \mathcal{V}$ to exactly one of the classes in $\mathcal{C}$.

\subsection{Graph Convolutional Networks}
We introduce the well-established multi-layer GCN~\cite{DBLP:conf/iclr/KipfW17} for node classification:
\begin{equation}
    H^{(l+1)} = \sigma(\Tilde{D}^{-\frac{1}{2}}\Tilde{A}\Tilde{D}^{-\frac{1}{2}}H^{(l)}W^{(l)}),\ l \geq 0,
\end{equation}
where $\Tilde{A}=A+I_N$ denotes the adjacency matrix with self-loop and the corresponding degree matrix is $\Tilde{D}$. $H^{(0)}=X$ and $\sigma$ is the activation function such as $\text{ReLU}$. For the $l$-th graph convolutional layer, we denote the node embeddings by $H^{(l)}$ and the learnable weight by $W^{(l)}$.

Without loss of generality, we consider a two-layer GCN with $\text{ReLU}$ activation in the hidden layer, which is commonly used in practice:
\begin{equation}
    Z=f_\theta(A,X) = \text{Softmax}(\hat{A}\cdot\text{ReLU}(\hat{A}XW^{(0)})W^{(1)}),
\end{equation}
where $\hat{A}=\Tilde{D}^{-\frac{1}{2}}\Tilde{A}\Tilde{D}^{-\frac{1}{2}}$. Let $f_\theta$ represent a GCN model with learnable parameters denoted as $\theta=\{W^{(0)}, W^{(1)}\}$. The optimal parameters $\theta$ are learned by minimizing cross-entropy on the output of the labeled nodes in $\mathcal{V}_\text{train}$:
\begin{equation}\label{eq:train}
    \mathcal{L}(f_\theta(A,X))=-\frac{1}{|\mathcal{V}_\text{train}|}\sum_{v\in \mathcal{V}_\text{train}} \text{ln}Z_{v, y_v}.
\end{equation}

\subsection{Adversarial Targeted Attack}
Below, we present the definition of adversarial targeted attack under the scenario of \textit{evasion} (test-time) attack. Note that it is straightforward to extend the definition to adversarial untargeted attacks~\cite{DBLP:conf/iclr/ZugnerG19} by suitably modifying the loss function.

\begin{definition}[\textbf{Adversarial targeted attack}]
    Given a graph $\mathcal{G}=(\mathcal{V},\mathcal{E})$, the goal of an attacker is to craft a perturbed graph $\mathcal{G}^\prime=(\mathcal{V}^\prime,\mathcal{E}^\prime)$ within a budget $\Delta$, and mislead the output of GCNs on a target node $u$. The adversarial attack on GCNs can be formulated as:
    \begin{equation}
        \nonumber
        \max_{\mathcal{G}^\prime \in \Phi(\mathcal{G})} \mathcal{L}_u(f_\theta(A^\prime,X)),
    \end{equation}
    where $A^\prime$ is the perturbed adjacency matrix that represents the graph $\mathcal{G}^\prime$, $\Phi(\mathcal{G})$ represents the set of all possible modified graphs that are constrained by the attack budget $\Delta$.
\end{definition}

Typically, an attacker aims to find a perturbed graph $\mathcal{G}^\prime$ that classifies target node $u$ as $y_u^\prime$ such that $y_u^\prime \neq y_u$, which is equivalent to maximizing the cross-entropy loss of the GCNs' output on $u$. According to~\cite{DBLP:conf/kdd/ZugnerAG18,li2021adversarial}, the perturbation can be performed on the targeted node $u$ or its neighborhoods, resulting in \textit{direct} attack or \textit{indirect/influence} attack, respectively. In this paper, we mainly consider the direct attack as it is more powerful than the indirect attack~\cite{chen2021understanding,DBLP:conf/kdd/ZugnerAG18,li2021adversarial}.

As an attacker usually has no access to the target model in a practical scenario, they instead train a surrogate model $f_{\theta^{*}}(A,X)$ locally and utilize the approximated loss $\mathcal{L}^{*} \approx \mathcal{L}$ to find the worst-case perturbations. The attack is also called \textit{surrogate attack}~\cite{DBLP:conf/kdd/ZugnerAG18}.

\subsection{Universal Defense on Graphs}
\label{sec:universal_defense}

We briefly outline necessary definitions of universal defense, a newly studied defense strategy on graphs. In this work, the term \textit{universal} means applicable to any \textit{node}~\cite{gua}, differs from that applicable to any \textit{image} in vision research~\cite{Moosavi-Dezfooli17}.

A universal defense mainly focuses on edge-injection attacks since (i) attackers tend to add edges between dissimilar nodes and (ii) injection is an operation with stronger attack power compared with deletion, the conclusions reached in prior studies~\cite{Wu0TDLZ19,chen2021understanding}. We provide further discussion in Section~\ref{sec:discuss}.
The core idea of universal defense is to generate a unique patch $p$ --- a length-$N$ binary vector $p$, where \texttt{1} indicates an attacker node\footnote{An attacker node is a maliciously added neighbor for a target node to achieve the adversarial goal~\cite{li2021adversarial}} and \texttt{0} otherwise. Without loss of generality, we denote these attacker nodes as $\mathcal{A}$ and term them as \textit{anchor nodes}, a set of important nodes that may be potentially used for attack. The universal patch $p$ is only computed once and can be applied to any node. When applied to a node $u$, the universal patch removes all potential adversarial edges (if exist) that are connected to the anchor nodes:
\begin{equation}
    \begin{aligned}
        A^\prime           & = A\circ (1-P) ,                                                                       \\
        \mathcal{E}^\prime & = \mathcal{E} - \{e=(u,v)\ |\ v \in \mathcal{A} \ \text{and}\ (u,v) \in \mathcal{E}\},
    \end{aligned}
\end{equation}
where $\circ$ is the element-wise product, $P \in \{0,1\}^{N \times N}$ is a derived matrix with the $u$-th row and $u$-th column replaced by the patch vector $p$. The element $(i,j)$ in $P$ equals \texttt{1} indicates the corresponding edge $(i,j) \in \mathcal{E}$ is to be removed (if exists). Let $\mathcal{G}_{(u)}=(\mathcal{V}, \mathcal{E}^\prime)$ denote the modified graph and $A^\prime$ the corresponding adjacency matrix \textit{w.r.t.} target node $u$, an effective defense should directly prune all of the malicious edges and mitigate the adversarial effects, while ensuring that the benign edges and downstream performance are not compromised.

\begin{figure}[t]
    \centering
    \includegraphics[width=\linewidth]{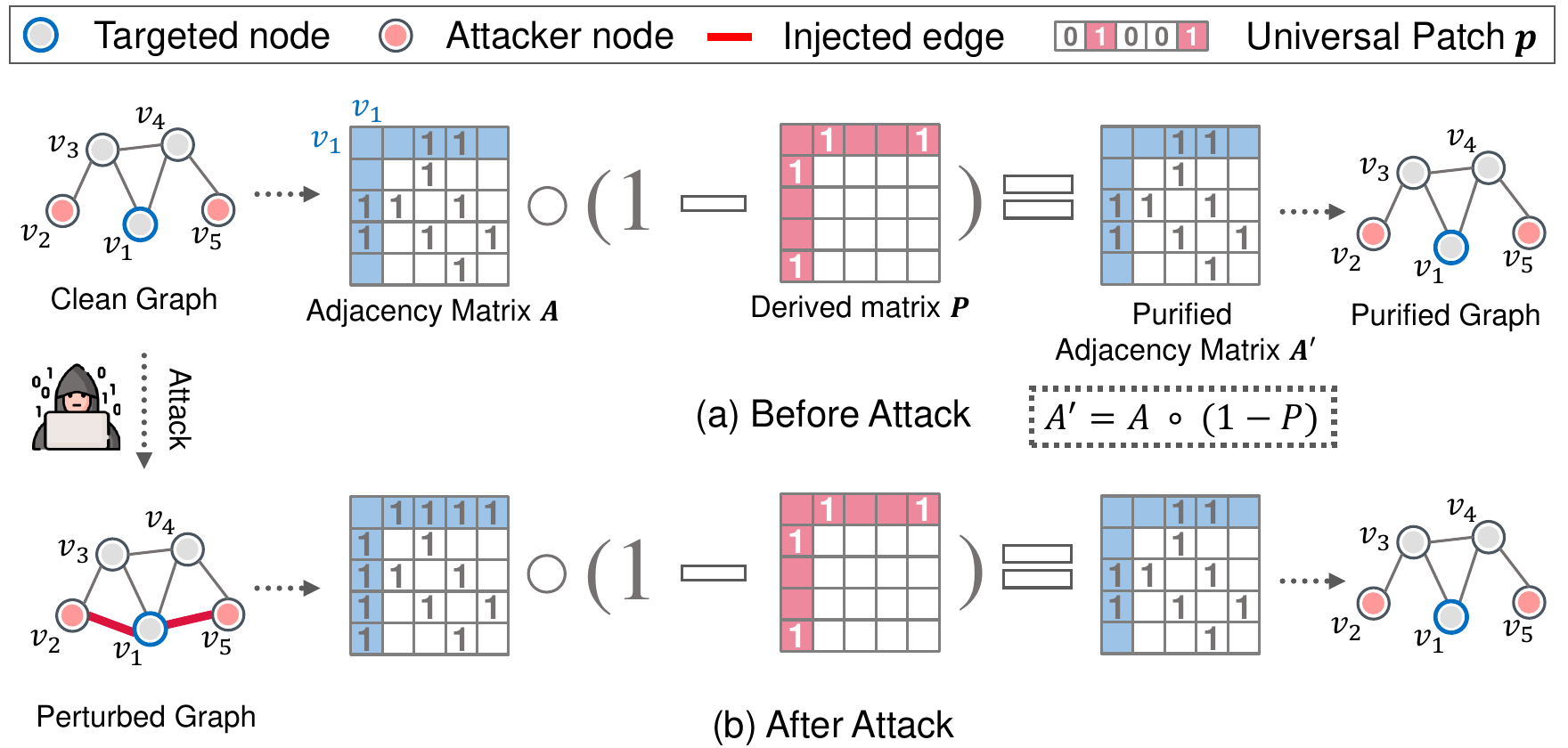}
    \caption{An illustrative example of graph universal defense. The universal patch $p$ can be applied to an arbitrary node (here $v_1$) to protect it from adversarial targeted attacks by removing adversarial edges (if exist).}
    \label{fig:universal_defense}
\end{figure}

Figure~\ref{fig:universal_defense} illustrates the universal defense in the context of adversarial targeted attacks. In this case, a targeted attack aims at fooling  GCNs on a specific node $v_1$ by modifying the graph structure. The universal defense, in contrast, generates a universal defensive patch $p$ which could be applied to any nodes to defend against adversarial attacks. We will elaborate in the next section an algorithm to find such $p$, or in other words, to identify the anchor nodes $\mathcal{A}$.

\section{Present Work}
In this section, we draw on the insights from the literature reviewed in Related Work, and empirically investigate the intrinsic patterns of adversarial attacks on graph data. Then, we analyze possible reasons to understand the observations and further introduce our proposed method.

\subsection{Empirical Investigation on Adversarial Attacks}
\label{sec:empirical_results}
In this subsection, we begin with an empirical study on Cora and PubMed (dataset statistics are listed in Table~\ref{tab:data}). Specifically, we attack different target nodes in the test set with three advanced surrogate attacks: SGA~\cite{li2021adversarial}, FGA~\cite{chen2018fast} and IG-FGSM (IG)~\cite{Wu0TDLZ19}, which craft the worst-case perturbations by leveraging the surrogate gradients in different ways. The perturbation budget for each target node is set as its degree, following \cite{li2021adversarial,DBLP:conf/kdd/ZugnerAG18}.

\begin{figure}[t]
    \centering
    \includegraphics[width=0.48\linewidth]{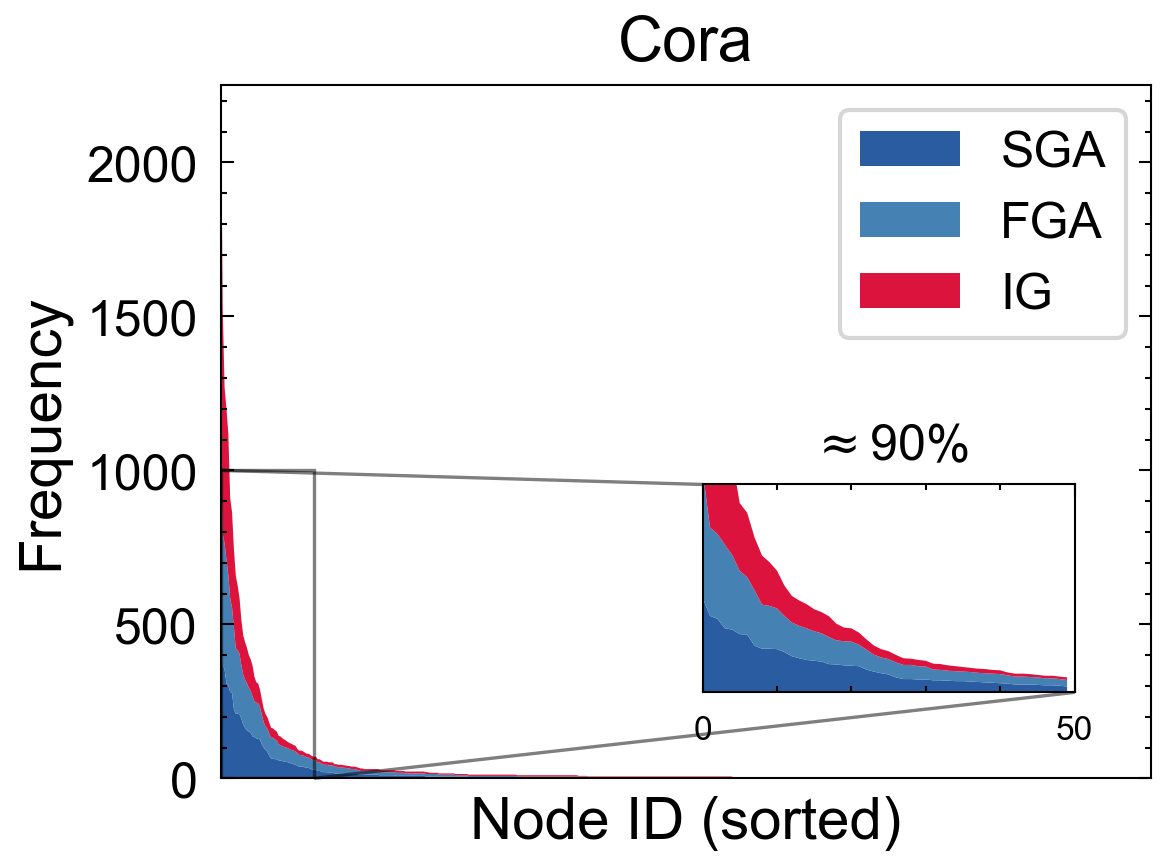}
    \includegraphics[width=0.48\linewidth]{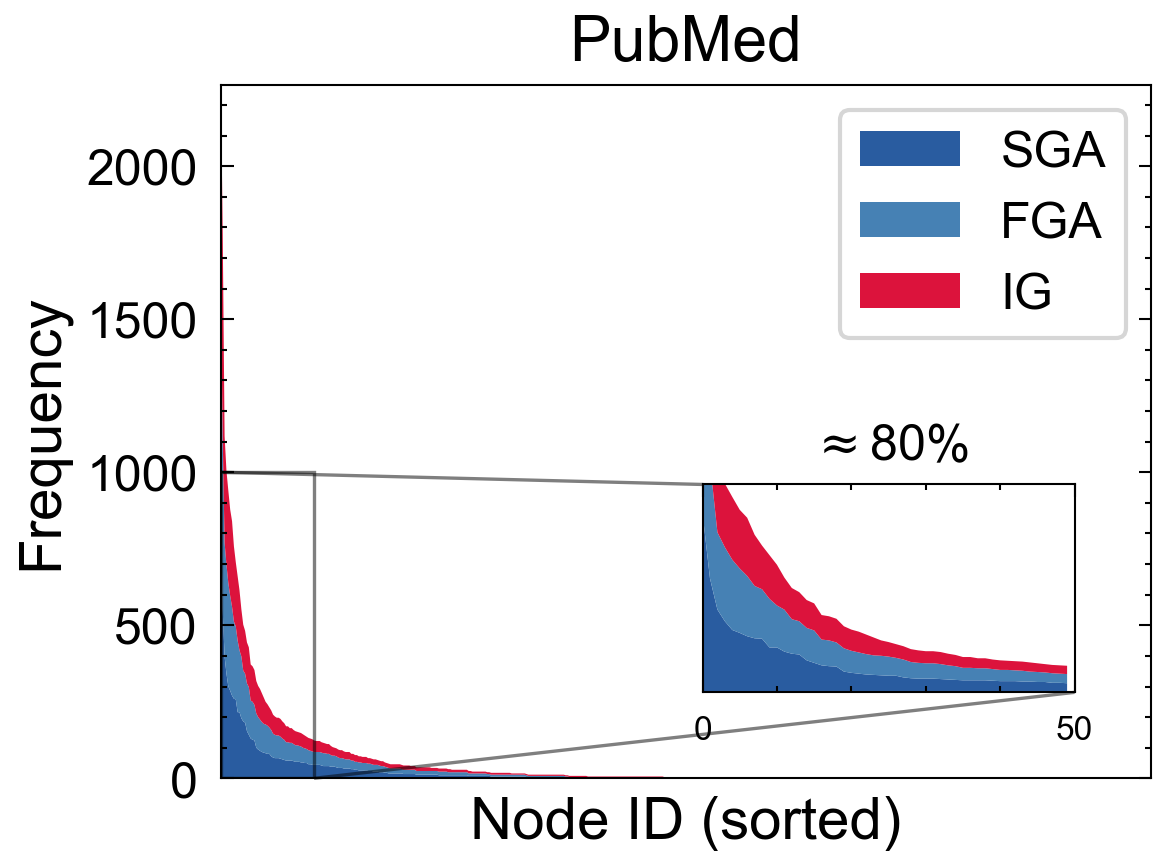}
    \caption{Frequency of top-500 selected attacker nodes by different attacks on Cora and PubMed datasets. The top-50 nodes account for almost 90\% and 80\% of the frequencies on both datasets, respectively.}
    \label{fig:distribution}
\end{figure}

We perturb different nodes with these attacks and count the frequency of each node selected as an attacker node, \textit{i.e.,} picked for adversarial edges.  We plot the frequency (in descending) of all nodes on both datasets in Figure \ref{fig:distribution}.

\textbf{Observation I.} {Attackers tend to connect the target node to a fixed set of attacker nodes, which also exhibits long-tailed distributions with a heavy imbalance in the measured frequency. As a result, the top-50 nodes with the highest frequencies account for nearly 90\% and 80\% of the frequencies on both datasets, respectively.}

\begin{figure}
    \centering
    \includegraphics[width=0.48\linewidth]{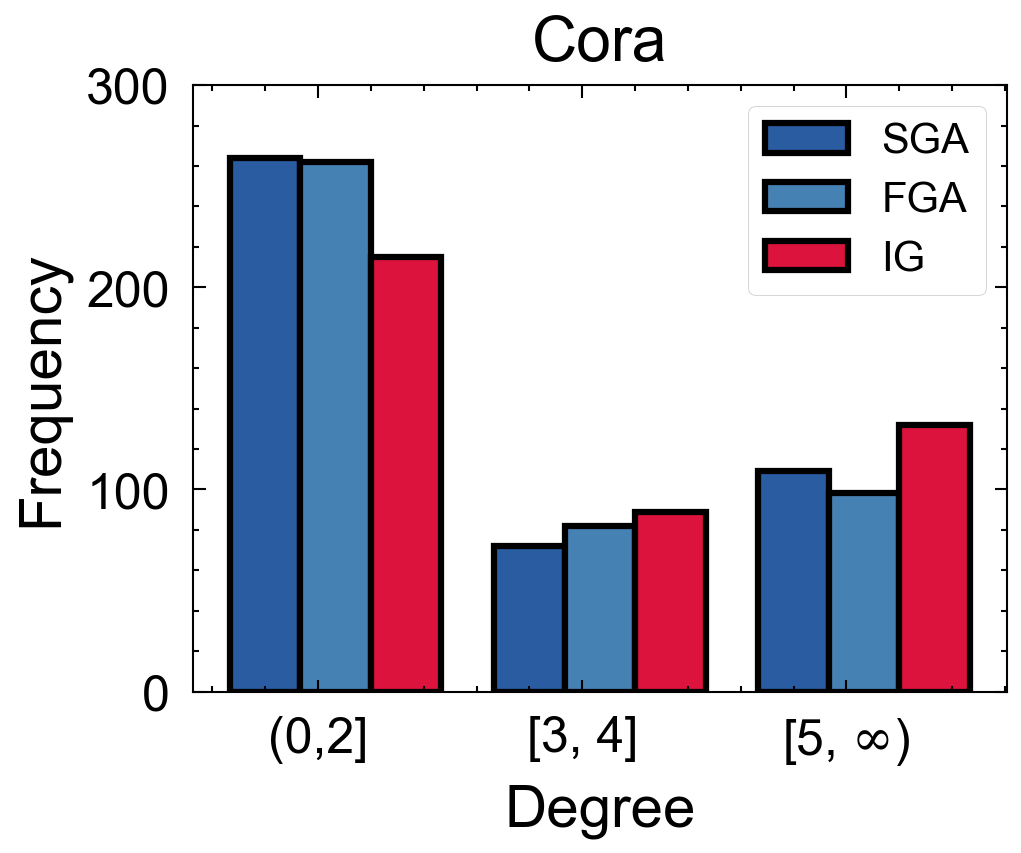}
    \includegraphics[width=0.48\linewidth]{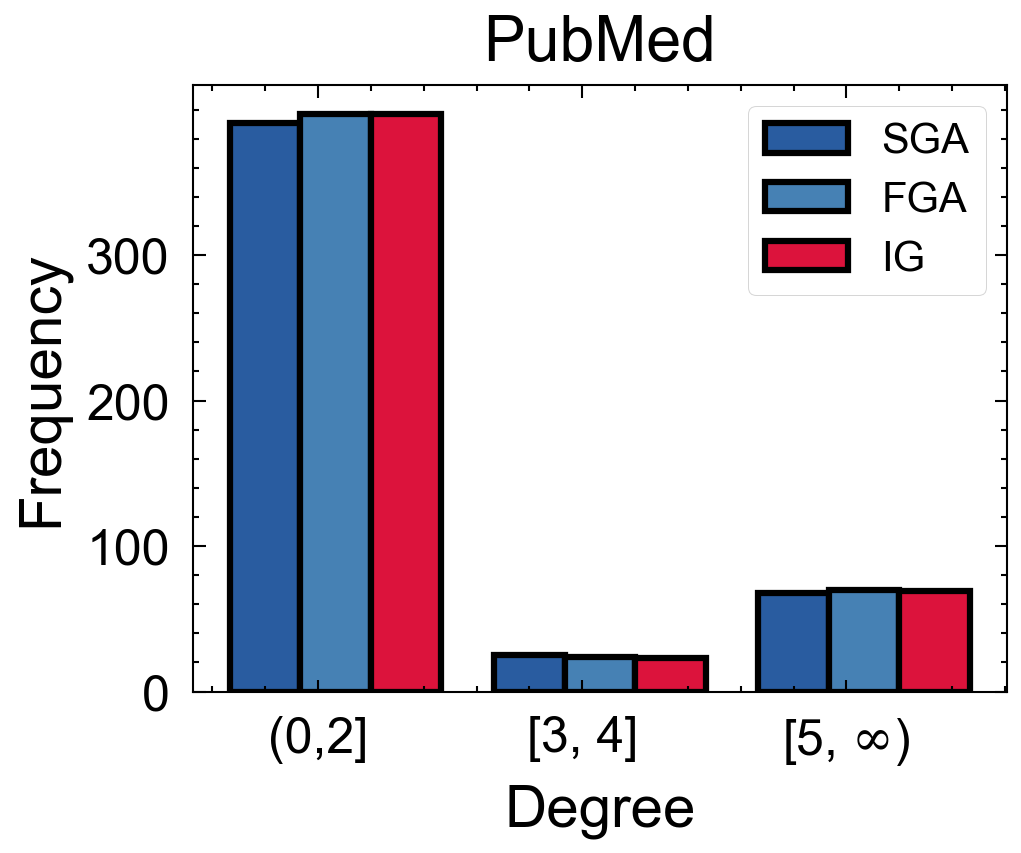}
    \caption{Degree distribution of top-500 selected attacker nodes by different attacks on Cora and PubMed datasets, respectively.}
    \label{fig:degree}
\end{figure}

Next, we plot the degree distribution of these high-frequency attacker nodes in Figure~\ref{fig:degree}.

\textbf{Observation II.} {Most of the attacker nodes are \textit{low-degree} nodes (\textit{i.e.}, degree $\leq 2$), and the phenomenon is more obvious on PubMed. The results suggest that adversarial edges tend to link the target node with low-degree nodes, and almost half of them are below 2 degrees. In other words, the low-degree nodes are more likely to be maliciously added neighbors for a target node. }

We will offer a further explanation on these findings in the next subsection.


\subsection{\textsc{Guard}: Universal Defense on Graphs}
\label{sec:guard}

The observation that a fixed set of low-degree nodes account for most of the measured frequencies is surprising. This gives immediately rise to a fundamental question: \textit{can we achieve a universal defense for an arbitrary node by uncovering these attacker nodes at test time?} In the following, we will address this question with our proposed \textsc{Guard}.

Recall that an optimal attack is typically achieved by exploiting the vulnerability of a locally trained surrogate model, which determines the (approximately) optimal perturbations by taking the gradient of the surrogate loss $\mathcal{L}^*$ \textit{w.r.t.} the adjacency matrix $A$. The largest magnitude of the gradient can be seen as a relaxation of the worst-case perturbation determined in the brute force method. Without loss of generality, we make the following assumption:
\begin{assumption}[\textbf{Optimal surrogate attack}]
    \label{assumption:attack}
    Given a target node $u$, the worst-case perturbation (modification) on $u$ is an edge $(u,v)$ with the largest magnitude of gradients $g_{u,v}$, where $g=[\frac{\partial{\mathcal{L}_u^{*}}}{\partial{A}}] \in \mathbb{R}^{N \times N}$ is the gradient matrix.
\end{assumption}

Assumption~\ref{assumption:attack} can be easily satisfied
and has been hold empirically in current works~\cite{li2021adversarial,Wu0TDLZ19,chen2018fast}. As we focus on the case of edge-injection attacks, the key to identifying attacker nodes is to find the edges corresponding to the maximum value of the gradients.

\begin{proposition}
    \label{proposition}
    For simplification, let us consider a 1-layer GCN with output $Z=\text{Softmax}(\hat{A}\mathcal{W})$ where $\mathcal{W}=XW$.
    Given a target node $u$ labeled as $y_u$, we have:
    \begin{equation}
        \nonumber
        (u,v^*)= {\arg \max}_{(u, v), v \notin \mathcal{N}(u)} \frac{\sum_{c \in \mathcal{C}} Z_{v,c}{\mathcal{W}_{v,c}}-\mathcal{W}_{v,y_u}}{\sqrt{d_v}},
    \end{equation}
    where $(u,v^*)$ denotes the edge corresponding to the largest gradient $g_{u, v^{*}}$ and $\mathcal{N}(u)$ is the set of nodes adjacent to $u$.
\end{proposition}

We give below the proof of Proposition~\ref{proposition}.
\begin{proof}
    The forward inference at the l-layer GCN is formally defined as:
    \begin{equation}
        Z=\text{Softmax}(\hat{A}\mathcal{W}), \quad \hat{A}=\Tilde{D}^{\frac{1}{2}}\Tilde{A}\Tilde{D}^{\frac{1}{2}}.
    \end{equation}
    For a target node $u$, attackers aim to find the worst-case perturbation with the approximated loss $\mathcal{L}_u^*=-\text{ln}Z_{u, y_u}$.
    Taking derivatives by applying the chain rule:
    \begin{equation}
        \begin{aligned}
            g=\frac{\partial{\mathcal{L}_u^*}}{\partial{A}} & =\frac{{\partial{\hat{A}}} }{\partial{A}}
            \frac{\partial{Z}_u}{\partial{\hat{A}}}  \frac{\partial{\mathcal{L}_u}}{\partial{Z}_u} \in \mathbb{R}^{N \times N}.
        \end{aligned}
    \end{equation}

    Following the chain rule of gradient backpropagation in terms of the adjacency matrix $A$, it is easy to calculate the gradient of each edge $(u,v)$ in the graph, or the element $(u,v)$ in $g$:
    \begin{equation}
        \label{eq:gradient}
        g_{u,v}=\frac{\hat{g}_{u,v}}{\sqrt{d_u d_v}}-0.5 \times \sum_{w \in \mathcal{N}(u)} \frac{d_w A_{u,w}}{(d_u d_w)^\frac{3}{2}}  (\hat{g}_{u,w}+\hat{g}_{w,u}),
    \end{equation}
    where $\hat{g}_{i,j}=[\partial{\mathcal{L}_u^*}/{\hat{A}}]_{i,j}=\sum_{c \in \mathcal{C}} Z_{i,c} \mathcal{W}_{j,c}-\mathcal{W}_{j,y_i}$.

    According to Eq.\eqref{eq:gradient}, the determining term of $g_{u,v}$ for different nodes $v$ is $\hat{g}_{u,v}/\sqrt{d_v}$ since $v \notin \mathcal{N}(u)$ and other terms are constants for a fixed target node $u$. For an edge $(u, v^*)$ with the largest value of gradient, we have:
    \begin{equation}
        \begin{aligned}
            (u, v^*) & ={\arg \max}_{(u, v), v \notin \mathcal{N}(u)} g_{u,v}                                                                                     \\
                     & ={\arg \max}_{(u, v), v \notin \mathcal{N}(u)} \frac{\hat{g}_{u,v}}{\sqrt{d_v}}                                                            \\
                     & ={\arg \max}_{(u, v), v \notin \mathcal{N}(u)} \frac{\sum_{c \in \mathcal{C}} Z_{v,c}{\mathcal{W}_{v,c}}-\mathcal{W}_{v,y_u}}{\sqrt{d_v}},
        \end{aligned}
    \end{equation}
    The desired result is attained.

\end{proof}

\begin{remark}\label{remark1}
    Essentially, it is shown in Proposition~\ref{proposition} that the magnitude of gradient $g_{u,v}$ is determined by the term $(\sum_{c \in \mathcal{C}} Z_{v,c}{\mathcal{W}_{v,c}}-\mathcal{W}_{v,y_u}) /\sqrt{d_v}$. That said, a node with a lower degree is more likely to be a maliciously added neighbor for a target node under Assumption~\ref{assumption:attack}.
\end{remark}

Proposition \ref{proposition} is basically built on a 1-layer GCN, we made this mild assumption to simplify our discussion and should not affect our findings. The result still holds for an $l$-layer GCN, where $\mathcal{W}$ is computed by collapsing weight matrices between consecutive layers, \textit{i.e.,} $\mathcal{W}=X \cdot W^{(0)}\cdots W^{(l-1)}$.

\begin{remark}\label{remark2}
    For any two target nodes $u_1$ and $u_2$, let $e_1^*=(u_1, v_1^*)$ and $e_2^*=(u_2, v_2^*)$ be two edges corresponding to the largest gradients for $u_1$ and $u_2$ respectively, then $v_1^*=v_2^*$ if $\mathcal{W}_{v_1,y_{u_1}} \approx \mathcal{W}_{v_2,y_{u_2}}$.
\end{remark}
The result is straightforward from Proposition~\ref{proposition}. The condition $\mathcal{W}_{v_1,y_{u_1}} \approx \mathcal{W}_{v_2,y_{u_2}}$ is often satisfied, since many modern neural networks are typically overconfident in their predictions~\cite{GuoPSW17}. That is, they often produce a high confidence probability for the predicted class, while treating all others equally with an equally low probability. The overconfidence issue of GCNs is also revealed in~\cite{li2021adversarial} and still holds for the linear part $\mathcal{W}$. In other words, attackers
would pick the same node when attacking different target nodes, which is in line with our empirical results in Figure~\ref{fig:distribution}.

Intuitively, for an edge to inject, we can determine how impactful that change was by looking at the gradient \textit{w.r.t.} the adjacency matrix $A$.
According to Proposition~\ref{proposition}, we measure the sensitivity of node $u$ to node $v$, or the influence of $v$ on $u$, by measuring the determining part of the gradient corresponding to the edge $(u,v)$. For a target node $u$, the influence score $\mathcal{I}_{u}(v)$ captures the relative influence of node $v$ on $u$, when $u$ being a target node:
\begin{equation}
    \label{eq:influence_original}
    \mathcal{I}_{u}(v) = \frac{\sum_{c \in \mathcal{C}} Z_{v,c}{\mathcal{W}_{v,c}}-\mathcal{W}_{v,y_u}}{d_v^\alpha},
\end{equation}
where $\alpha$ is a scaling factor that controls the impact of node degree and $\alpha=0.5$ for a standard case in Proposition~\ref{proposition}. $\mathcal{W}$ can be easily obtained by training a surrogate SGC or linear GCN locally. Note that $\sum_{c \in \mathcal{C}} Z_{v,c}=1$, we use the upper bound of $\mathcal{I}_{u}(v)$ as the approximated influence score for the ease of computation:
\begin{equation}
    \mathcal{I}_{u}^*(v) = \frac{\max_{c \in \mathcal{C}}{\mathcal{W}_{v,c} - \mathcal{W}_{v,y_u}}}{d_v^\alpha}.
\end{equation}
By using the upper bound of $\mathcal{I}_u(v)$, we can hereby reduce the computational complexity of our approach while still achieving a reasonable approximation of the true influence score.

\begin{algorithm}[t]
    \caption{Graph Universal Adversarial Defense}
    \label{algo:guard}
    \begin{algorithmic}[0]
        \Require Graph $\mathcal{G}=(\mathcal{V}, \mathcal{E})$; node features $X$ and degrees $d$; labeled nodes set $\mathcal{V}_\text{sub}$, weight matrix $W$, target node $u$, hyperparameters $k$ and $\alpha$;
        \Ensure Purified Graph $\mathcal{G}_{(u)}=(\mathcal{V}, \mathcal{E}^\prime)$ for target node $u$;
    \end{algorithmic}
    \begin{algorithmic}[1]
        \State $\mathcal{W}\gets X \cdot W$;
        \State $\mathcal{I}^*(v) \gets 0,\ \forall v \in \mathcal{V}$; \Comment{Initialize node influence score.}
        \For {$v \in \mathcal{V}$}
        \State $\mathcal{I}^*(v) \gets \frac{1}{d_v^\alpha}(\max_{c \in \mathcal{C}}{\mathcal{W}_{v,c} - \frac{1}{|\mathcal{V}_\text{sub}|} \sum_{s \in \mathcal{V}_\text{sub}}  \mathcal{W}_{v,y_s}})$;
        \EndFor
        \State $\mathcal{A} \gets \{v\ |\ \mathcal{I}^*(v)\ \text{is}\ k$-$\text{largest}\}$; \Comment{Anchor nodes.}
        \State $\mathcal{E}^\prime \gets \mathcal{E} - \{e=(u,v)\ |\ v \in \mathcal{A} \ \text{and}\ (u,v) \in \mathcal{E}\}$;\\
        \Return $\mathcal{G}_{(u)}=(\mathcal{V}, \mathcal{E}^\prime)$;
    \end{algorithmic}
\end{algorithm}

To simulate the attacks on different target nodes $u$, we can compute the influence distribution of all nodes based on a subset of labeled nodes $\mathcal{V}_{\text{sub}} \subseteq \mathcal{V}_\text{train}$, by averaging the influence score $\mathcal{I}_u(v)$ \textit{w.r.t.} different target node $u \in \mathcal{V}_{\text{sub}}$:
\begin{equation}
    \label{eq:influence}
    \begin{aligned}
        \mathcal{I}^*(v) & = \frac{1}{|\mathcal{V}_\text{sub}|}\sum_{u \in \mathcal{V}_\text{sub}}\mathcal{I}_{u}^*(v)                                                                      \\
                         & =\frac{1}{d_v^\alpha}(\max_{c \in \mathcal{C}}{\mathcal{W}_{v,c} - \frac{1}{|\mathcal{V}_\text{sub}|} \sum_{u \in \mathcal{V}_\text{sub}}  \mathcal{W}_{v,y_u}})
    \end{aligned}
\end{equation}

Under Assumption~\ref{assumption:attack}, the node with the highest influence score is the most likely to be an attacker node. Therefore, we derive the set of anchor nodes (size $k$) as:
\begin{equation}
    \mathcal{A} = \{v\ |\ \mathcal{I}^*(v)\ \text{is}\ k\text{-largest}\}.
\end{equation}
There is a trade-off between performance and robustness, with a larger $k$ would result in a better defense, but might sacrifice predictive accuracy on clean graphs as it removes a large proportion of edges. The detailed algorithm of \textsc{Guard} is described in Algorithm~\ref{algo:guard}.

\section{Discussion}
\label{sec:discuss}
In this section, we provide further discussion of our proposed \textsc{Guard}. We first explain the reason why \textsc{Guard} focuses only on edge injection attacks rather than edge deletion attacks, and then discuss its time complexity.

\subsection{Edge Injection Attack \& Deletion Attack}
We focus on edge insertion rather than edge deletion due to the following reasons: (i) \textbf{Attackers tend to add edges between dissimilar nodes.} This is an important conclusion reached in \cite{Wu0TDLZ19}, and also validated in \cite{jin2020graph}. Therefore, a simple similarity-based preprocessing method achieves is able to achieve good defensive results.
(ii) \textbf{Insertion is a more powerful attack operation compared to compared with deletion.} In~\cite{chen2021understanding}, the authors conduct an empirical study by enumerating all possible attacks of one-edge insertion and one-edge deletion. The results show that one-edge insertion attacks are significantly more successful in attacking nodes than one-edge deletion attacks. These findings are consistent with the previous study \cite{DBLP:conf/nips/BojchevskiG19}.
(iii) \textbf{Edge insertion attacks are more practical for real-world scenarios.} In practice, attackers may not have enough access to execute the `deletion' manipulation due to legal restrictions. For example, in a recommender system, users can only buy/add new items (edge insertion) instead of deleting their purchase/interaction histories (edge deletion).
(iv) \textbf{Edge deletion attack is not usually the main concern for the vulnerability of GNNs.} Even in an extreme case, \textit{i.e.}, an attacker removes all the edges connected to the target node, the GCNs are at worst degraded to MLPs. Nevertheless, MLPs are also able to maintain a certain classification accuracy although slightly underperforming GCNs.

\subsection{Time Complexity}
Here we discuss the time complexity of \textsc{Guard}. As described above, the overall time complexity of \textsc{Guard} is $\mathcal{O}(N|\mathcal{V}_\text{sub}|+N\log k)$ \footnote{Note that $|\mathcal{V}_\text{sub}|$ and $k$ are often small.}. Specifically, the computation of \textsc{Guard} consists of three main steps: (i) obtain $\mathcal{W}=X \cdot W^{(0)}\cdots W^{(l-1)}$; (ii) compute the approximated influence score $\mathcal{I}_v^*$ for each node $v \in \mathcal{V}$; (iii) identify $k$ nodes with largest $\mathcal{I}^*$ as the anchor nodes $\mathcal{A}$. Note that, step (i) is not usually the major bottleneck of \textsc{Guard}, since $\mathcal{W}$ can be obtained directly from the victim GCNs. Taken together, the overall time complexity of \textsc{Guard} is $\mathcal{O}(N|\mathcal{V}_\text{sub}|+N\log k))$, where $N$ is the number of nodes, $\mathcal{V}_\text{sub}$ is a subset of labeled nodes, $k$ is the number of anchor nodes. The computation of $\mathcal{I}^*$ in step (ii) can be trivially parallelized, and $k$ is usually small ($k \ll N$), thus the overall computation overhead is low and acceptable. Besides, the computation needs to be performed only once for any node in the graph, which also benefits the scalability of \textsc{Guard}.

\begin{table}[h]
    \centering
    \caption{Dataset Statistics. For Cora and PubMed we extract the largest connected component of graphs.}\label{tab:data}
    \resizebox{\linewidth}{!}{\begin{tabular}{l|cccc}
            \toprule
            \textbf{}     & \textbf{Cora} & \textbf{PubMed} & \textbf{arXiv} & \textbf{Reddit} \\
            \midrule
            \# Nodes      & 2,485         & 19,717          & 169,343        & 232,965         \\
            \# Edges      & 10,138        & 88,648          & 1,166,243      & 11,606,919      \\
            \# Features   & 1,433         & 500             & 128            & 602             \\
            \# Classes    & 7             & 3               & 6              & 41              \\
            Feature Type  & Binary        & Binary          & Continuous     & Continuous      \\
            Avg. degree   & 4.08          & 4.50            & 13.70          & 99.65           \\
            degree$\leq$2 & 36\%          & 63\%            & 65\%           & 3\%             \\
            \bottomrule
        \end{tabular}
    }
\end{table}

\section{Experiments}
In this section, we perform experimental evaluations of our proposed \textsc{Guard} method.
The goal of our experiments is to test whether GCNs armed with \textsc{Guard} are more robust and reliable against adversarial targeted attacks, particularly compared with current defenses.
In what follows, we first introduce the experimental settings and then present empirical results.

\subsection{Experimental Settings}

\begin{table*}[t]
    \centering
    \caption{Comparison of clean accuracy (\%) with preprocessing-based defenses. (OOM: out of memory. N/A: not applicable.)}\label{tab:clean}
    \resizebox{\linewidth}{!}
    {\begin{tabular}{l|cccccc|cccccc}

            \toprule
                            & GCN            & +Jaccard             & +SVD                 & +\textsc{Rand} & +\textsc{Deg}  & +\textsc{Guard} & SGC            & +Jaccard             & +SVD                 & +\textsc{Rand} & +\textsc{Deg}  & +\textsc{Guard} \\
            \midrule
            \textbf{Cora}   & 82.7${\pm0.5}$ & 81.6${\pm0.7}$       & 77.8${\pm0.9}$       & 82.7${\pm0.4}$ & 82.7${\pm0.2}$ & 82.7${\pm0.3}$  & 83.1${\pm0.8}$ & 82.5${\pm0.6}$       & 77.6${\pm1.0}$       & 82.6${\pm0.5}$ & 83.3${\pm0.3}$ & 83.1${\pm0.5}$  \\
            \textbf{PubMed} & 84.2${\pm0.3}$ & 84.2${\pm0.6}$       & \colorbox{gray}{OOM} & 84.1${\pm0.2}$ & 84.2${\pm0.1}$ & 84.3${\pm0.4}$  & 83.1${\pm0.4}$ & 83.3${\pm0.5}$       & \colorbox{gray}{OOM} & 82.8${\pm0.5}$ & 83.1${\pm0.3}$ & 83.3${\pm0.3}$  \\
            \textbf{arXiv}  & 65.1${\pm0.5}$ & \colorbox{gray}{N/A} & \colorbox{gray}{OOM} & 64.5${\pm0.2}$ & 65.1${\pm0.3}$ & 65.0${\pm0.2}$  & 63.4${\pm0.7}$ & \colorbox{gray}{N/A} & \colorbox{gray}{OOM} & 63.1${\pm0.5}$ & 63.4${\pm0.5}$ & 63.3${\pm0.4}$  \\
            \textbf{Reddit} & 92.7${\pm0.5}$ & \colorbox{gray}{N/A} & \colorbox{gray}{OOM} & 92.4${\pm0.2}$ & 92.9${\pm0.2}$ & 92.7${\pm0.4}$  & 94.3${\pm0.6}$ & \colorbox{gray}{N/A} & \colorbox{gray}{OOM} & 94.0${\pm0.3}$ & 94.1${\pm0.4}$ & 94.3${\pm0.2}$  \\
            \bottomrule
        \end{tabular}
    }

\end{table*}

\begin{table*}[t]
    \centering
    \caption{Comparison of classification accuracy (\%) with \textit{preprocessing-based} defenses against adversarial targeted attacks. The best results on GCN and SGC are \textbf{boldfaced}, respectively. }\label{tab:after_attack}
    \resizebox{\linewidth}{!}{
        \begin{tabular}{ll|cccccc|cccccc}

            \toprule
            \textbf{Dataset} & \textbf{Attack}
                             & \textbf{GCN}    & \textbf{+Jaccard}    & \textbf{+SVD}           & \textbf{\textsc{+Rand}} & \textbf{\textsc{+Deg}} & \textbf{\textsc{+Guard}}
                             & \textbf{SGC}    & \textbf{+Jaccard}    & \textbf{+SVD}           & \textbf{\textsc{+Rand}} & \textbf{\textsc{+Deg}} & \textbf{\textsc{+Guard}}                                                                                                                                                        \\
            \midrule

            \multirow{5}{*}{\textbf{Cora}}
                             & {SGA}
                             & 13.8${\pm0.4}$  & 30.1${\pm0.4}$       & 69.5${\pm0.7}$          & 15.3${\pm0.4}$          & 38.7${\pm0.2}$         & \textbf{81.7${\pm0.3}$}
                             & 3.2${\pm0.3}$   & 24.3${\pm0.5}$       & 70.0${\pm0.6}$          & 5.3${\pm0.2}$           & 36.3${\pm0.3}$         & \textbf{81.4${\pm0.2}$}                                                                                                                                                         \\
                             & {FGA}
                             & 11.9 ${\pm0.3}$ & 31.7${\pm0.5}$       & 74.2${\pm0.6}$          & 14.2${\pm0.4}$          & 34.9${\pm0.3}$         & \textbf{75.5${\pm0.4}$}
                             & 13.4${\pm0.6}$  & 34.6${\pm0.6}$       & 74.9${\pm0.9}$          & 16.8${\pm0.5}$          & 40.2${\pm0.4}$         & \textbf{79.9${\pm0.5}$}                                                                                                                                                         \\
                             & {IG}
                             & 15.2${\pm0.8}$  & 40.9${\pm0.6}$       & \textbf{77.5${\pm0.8}$} & 16.1${\pm0.4}$          & 29.5${\pm0.6}$         & 64.8${\pm0.6}$
                             & 12.9${\pm0.7}$  & 44.4${\pm0.7}$       & \textbf{75.8${\pm1.2}$} & 15.3${\pm0.5}$          & 37.1${\pm0.6}$         & 72.9${\pm0.6}$                                                                                                                                                                  \\
                             & {RBCD}          & 10.7${\pm0.3}$       & 28.8${\pm0.6}$          & 70.1${\pm0.4}$          & 14.5${\pm0.8}$         & 35.2                     & \textbf{71.3${\pm0.4}$} & 8.5${\pm0.5}$ & 23.9${\pm0.5}$       & 69.4${\pm0.6}$          & 12.5${\pm0.5}$ & 39.6${\pm0.3}$ & \textbf{78.7${\pm0.7}$} \\
                             & {Nettack}       & 7.9${\pm0.8}$        & 35.4${\pm0.6}$          & \textbf{65.2${\pm0.3}$} & 10.9${\pm0.7}$         & 22.6${\pm0.9}$           & 56.4${\pm0.2}$          & 4.8${\pm0.6}$ & 19.2${\pm0.4}$       & \textbf{62.7${\pm0.4}$} & 8.4${\pm0.3}$  & 22.7${\pm0.5}$ & 59.4${\pm0.2}$          \\
            \midrule

            \multirow{5}{*}{\textbf{PubMed}}
                             & {SGA}
                             & 4.0${\pm0.4}$   & 8.6${\pm0.2}$        & \colorbox{gray}{OOM}    & 4.3${\pm0.3}$           & 4.0${\pm0.3}$          & \textbf{77.6${\pm0.2}$}
                             & 1.3${\pm0.3}$   & 6.4${\pm0.3}$        & \colorbox{gray}{OOM}    & 1.9${\pm0.5}$           & 1.3${\pm0.2}$          & \textbf{76.7${\pm0.2}$}                                                                                                                                                         \\
                             & {FGA}
                             & 2.6${\pm0.4}$   & 5.9${\pm0.5}$        & \colorbox{gray}{OOM}    & 2.9${\pm0.5}$           & 2.6${\pm0.4}$          & \textbf{72.7${\pm0.4}$}
                             & 3.7${\pm0.6}$   & 6.5${\pm0.4}$        & \colorbox{gray}{OOM}    & 4.1${\pm0.6}$           & 3.7${\pm0.4}$          & \textbf{71.5${\pm0.5}$}                                                                                                                                                         \\
                             & {IG}
                             & 9.6${\pm0.5}$   & 14.1${\pm0.4}$       & \colorbox{gray}{OOM}    & 9.9${\pm0.4}$           & 9.6${\pm0.4}$          & \textbf{77.1${\pm0.3}$}
                             & 9.5${\pm0.4}$   & 14.0${\pm0.3}$       & \colorbox{gray}{OOM}    & 10.3${\pm0.6}$          & 9.5${\pm0.2}$          & \textbf{76.8${\pm0.3}$}                                                                                                                                                         \\
                             & {RBCD}          & 2.3${\pm0.8}$        & 7.4${\pm0.4}$           & \colorbox{gray}{OOM}    & 8.4${\pm0.7}$          & 10.2${\pm0.7}$           & \textbf{71.9${\pm0.2}$} & 2.1${\pm0.5}$ & 6.3${\pm0.6}$        & \colorbox{gray}{OOM}    & 8.8${\pm0.5}$  & 9.5${\pm0.8}$  & \textbf{70.4${\pm0.3}$} \\
            \midrule

            \multirow{2}{*}{\textbf{arXiv}}
                             & {SGA}
                             & 0.5${\pm0.5}$   & \colorbox{gray}{N/A} & \colorbox{gray}{OOM}    & 1.1${\pm0.5}$           & 10.1${\pm0.6}$         & \textbf{62.3${\pm0.3}$}
                             & 0.3${\pm0.9}$   & \colorbox{gray}{N/A} & \colorbox{gray}{OOM}    & 0.9${\pm0.9}$           & 10.4${\pm0.8}$         & \textbf{62.0${\pm0.8}$}                                                                                                                                                         \\
                             & {RBCD}          & 0.3${\pm0.1}$        & \colorbox{gray}{N/A}    & \colorbox{gray}{OOM}    & 0.8${\pm0.4}$          & 6.4${\pm0.3}$            & \textbf{60.9${\pm0.5}$} & 0.1${\pm0.2}$ & \colorbox{gray}{N/A} & \colorbox{gray}{OOM}    & 0.6${\pm0.6}$  & 9.2${\pm0.7}$  & \textbf{61.7${\pm0.4}$} \\
            \midrule

            \multirow{2}{*}{\textbf{Reddit}}
                             & {SGA}
                             & 3.1${\pm0.2}$   & \colorbox{gray}{N/A} & \colorbox{gray}{OOM}    & 4.2${\pm0.3}$           & 86.5${\pm0.6}$         & \textbf{89.8${\pm0.4}$}
                             & 0.0${\pm0.0}$   & \colorbox{gray}{N/A} & \colorbox{gray}{OOM}    & 0.0${\pm0.1}$           & 84.1${\pm0.7}$         & \textbf{87.2${\pm0.3}$}                                                                                                                                                         \\
                             & {RBCD}          & 2.7${\pm0.1}$        & \colorbox{gray}{N/A}    & \colorbox{gray}{OOM}    & 3.8${\pm0.5}$          & 72.4${\pm0.7}$           & \textbf{87.9${\pm0.6}$} & 0.3${\pm0.0}$ & \colorbox{gray}{N/A} & \colorbox{gray}{OOM}    & 0.3${\pm0.0}$  & 74.5${\pm0.5}$ & \textbf{85.6${\pm0.8}$} \\
            \bottomrule
        \end{tabular}
    }

\end{table*}

\subsubsection{Datasets.}
The experiments are conducted on four real-world benchmark datasets, including three citation networks, \textit{i.e.}, Cora, PubMed~\cite{DBLP:journals/aim/SenNBGGE08}, and ogbn-arXiv (arXiv)~\cite{ogb}, one social graph dataset Reddit~\cite{hamilton2017inductive}. For Cora and PubMed, we preprocess and split them the same as~\cite{li2021adversarial,chen2021understanding}, which extract the largest connected component of the graph and split the dataset into 10\%/10\%/80\% for training/validation/testing. For arXiv and Reddit, we use public splits and settings in our experiments. Statistics of these datasets are summarized in Table~\ref{tab:data}.

\subsubsection{Attacks.}
We employ five state-of-the-art adversarial targeted attacks, including four gradient-based attacks, \textit{i.e.}, FGA~\cite{chen2018fast}, IG (short for IG-FGSM~\cite{Wu0TDLZ19}), SGA~\cite{li2021adversarial}, and RBCD~\cite{rbcd}, and one greedy-based attack Nettack~\cite{DBLP:conf/kdd/ZugnerAG18}. All these methods take GCN or SGC as surrogate models to conduct transfer attacks. Following~\cite{li2021adversarial}, we randomly choose 1,000 nodes from the test set as target nodes for each dataset. We define the perturbation budget $\Delta=d_u$ for a target node $u$ where $d_u$ is the degree of node $u$, as advocated in~\cite{li2021adversarial,DBLP:conf/kdd/ZugnerAG18}.

\subsubsection{Defenses.}
As graph universal defenses were barely studied, we design two relevant baselines for \textsc{Guard}, including \textsc{Rand} and \textsc{Deg}. \textsc{Rand} picks anchor nodes randomly while \textsc{Deg} picks anchors nodes with the lowest degrees. Additionally, we compare \textsc{Guard} with other non-universal defense methods: Jaccard~\cite{Wu0TDLZ19}, SVD~\cite{EntezariADP20}, RGCN~\cite{DBLP:conf/kdd/ZhuZ0019}, SimPGCN~\cite{simpgcn}, ElasticGNN~\cite{liu2021elastic}, MedianGCN~\cite{chen2021understanding},
SoftMedian~\cite{rbcd}, and GNNGUARD~\cite{gnnguard}. Among these methods, SVD and Jaccard are \textit{preprocessing-based} methods, which filter adversarial perturbations based on a (dense) low-rank approximation of the adjacency matrix and feature dissimilarity, respectively. They are not universal defenses.
RGCN, SimPGCN, ElasticGNN, MedianGCN, SoftMedian, and GNNGUARD are improved GCNs (\textit{model-based}) with robust architectures, message passing schemes, or regularizations. All the defense methods are configured according to the best performance setting in their results.

\subsubsection{Victim models.} In our evaluation, we use GCN~\cite{DBLP:conf/iclr/KipfW17} and SGC~\cite{WuSZFYW19} as the default victim models. We use two-layer GCN with hidden units 16 for Cora, PubMed, and Reddit, and three-layer GCN with hidden units 256 for arXiv. For SGC, the number of layers is set as 2 across all datasets. We train all models for 200 epochs using Adam~\cite{adam} optimizer, with an initial learning rate of 0.01. The best models are picked according to their performance on the validation set.

\subsubsection{Hyperparameter setting.}  For universal defense methods, \textit{i.e.}, \textsc{Rand}, \textsc{Deg} and \textsc{Guard}, we set different $k$ across different datasets, where $k=200$ for Cora, $k=500$ PubMed, $k=10,000$ for arXiv and $k=20,000$ for Reddit. The scaling factor $\alpha$ for \textsc{Guard} is set to $2$ across all datasets. We discuss them in Section~\ref{sec:explor}.
For baseline methods, the approximated rank of SVD is set as 50 and the threshold of Jaccard is set as 0.01 to filter adversarial perturbations. For the remaining configuration, we closely follow the setup of~\cite{chen2021understanding,li2021adversarial}.

\subsubsection{Evaluation protocol.} We first use various attacks to obtain the perturbed graphs \textit{w.r.t.} the 1,000 target nodes in test set. Then, for all generated graphs we record the classification accuracy of GCNs on these target nodes. We evaluate the models/methods on the evasive setting, \textit{i.e.}, the attack happens after the model is trained. Performance is reported by the average accuracy with standard deviation based on five runs on the clean/perturbed graphs.

\subsubsection{Implementation details.}
We implement our method in PyTorch~\cite{pytorch} and DGL~\cite{dgl}. For the other methods,
we use all the original papers' code from their GitHub pages.
All experiments are conducted on an NVIDIA RTX 3090 Ti GPU with 24 GB memory unless specified.
Code for reproducibility is available at \url{https://github.com/EdisonLeeeee/GUARD}.

\begin{table*}[t]
    \centering
    \caption{Comparison of classification accuracy (\%) with \textit{model-based} defenses against adversarial targeted attacks.}\label{tab:after_attack2}
    \resizebox{\linewidth}{!}
    {
        \begin{tabular}{l|ccccc|cccc|cc|cc}
            \toprule
            \multirow{2}{*}{\textbf{Method}} & \multicolumn{5}{c}{\textbf{Cora}} & \multicolumn{4}{c}{\textbf{PubMed}} & \multicolumn{2}{c}{\textbf{arXiv}} & \multicolumn{2}{c}{\textbf{Reddit}}                                                                                                                                    \\
            \cmidrule{2-14}

                                             & SGA                               & FGA                                 & IG                                 & RBCD                                & Nettack                 & SGA                     & FGA                  & IG                   & RBCD & SGA & RBCD & SGA & RBCD \\
            \midrule

            GCN                              & 13.8${\pm0.4}$                    & 11.9${\pm0.3}$                      & 15.2${\pm0.8}$                     & 10.7${\pm0.3}$                      & 8.7${\pm0.9}$
                                             & 4.0${\pm0.4}$                     & 2.6${\pm0.4}$                       & 9.6${\pm0.5}$                      & 2.3${\pm0.8}$                       & 0.5${\pm0.5}$           & 0.3${\pm0.1}$           & 3.1${\pm0.2}$        & 2.7${\pm0.1}$                                         \\
            SGC                              & 3.2${\pm0.3}$                     & 13.4${\pm0.6}$                      & 12.9${\pm0.7}$                     & 8.5${\pm0.8}$                       & 4.8${\pm0.6}$
                                             & 1.3${\pm0.3}$                     & 3.7${\pm0.6}$                       & 9.5${\pm0.4}$                      & 2.1${\pm0.5}$                       & 0.3${\pm0.9}$           & 0.1${\pm0.2}$           & 0.0${\pm0.0}$        & 0.3${\pm0.0}$                                         \\
            \midrule
            RGCN                             & 21.4${\pm0.4}$                    & 15.4${\pm0.5}$                      & 15.4${\pm0.6}$                     & 12.2${\pm0.7}$                      & 9.1${\pm0.9}$
                                             & 12.4${\pm0.2}$                    & 13.6${\pm0.5}$                      & 19.0${\pm0.4}$                     & 4.9${\pm0.5}$                       & \colorbox{gray}{OOM}    & \colorbox{gray}{OOM}    & \colorbox{gray}{OOM} & \colorbox{gray}{OOM}                                  \\
            SimPGCN                          & 11.4${\pm0.5}$                    & 13.5${\pm0.6}$                      & 12.5${\pm0.3}$                     & 14.2${\pm0.3}$                      & 8.3${\pm0.7}$
                                             & 16.4${\pm0.5}$                    & 13.5${\pm0.5}$                      & 18.6${\pm0.7}$                     & 9.4${\pm0.9}$                       & \colorbox{gray}{OOM}    & \colorbox{gray}{OOM}    & \colorbox{gray}{OOM} & \colorbox{gray}{OOM}                                  \\
            ElasticGNN                       & 35.5${\pm0.4}$                    & 37.2${\pm0.5}$                      & 35.8${\pm0.4}$                     & 32.7${\pm0.4}$                      & 29.5${\pm0.2}$
                                             & 22.6${\pm0.3}$                    & 20.4${\pm0.4}$                      & 27.2${\pm0.2}$                     & 23.3${\pm0.8}$                      & \colorbox{gray}{OOM}    & \colorbox{gray}{OOM}    & \colorbox{gray}{OOM} & \colorbox{gray}{OOM}                                  \\
            MedianGCN
                                             & 34.9${\pm0.7}$                    & 35.1${\pm0.6}$                      & 42.3${\pm0.8}$                     & 32.5${\pm0.5}$                      & 31.4${\pm0.7}$
                                             & 52.3${\pm0.4}$                    & 50.2${\pm0.7}$                      & 59.4${\pm0.7}$                     & 51.7${\pm0.8}$                      & \colorbox{gray}{OOM}    & \colorbox{gray}{OOM}    & \colorbox{gray}{OOM} & \colorbox{gray}{OOM}                                  \\
            SoftMedian
                                             & 35.2${\pm0.3}$                    & 36.4${\pm0.3}$                      & 45.5${\pm0.6}$                     & 33.7${\pm0.6}$                      & 32.9${\pm0.1}$
                                             & 51.6${\pm0.2}$                    & 54.7${\pm0.5}$                      & 61.4${\pm0.8}$                     & 55.3${\pm0.4}$                      & 53.2${\pm0.8}$          & 50.7${\pm0.6}$          & 78.6${\pm0.9}$       & 74.3${\pm0.8}$                                        \\
            GNNGUARD                         & 59.9${\pm0.3}$                    & 54.1${\pm0.3}$                      & 62.6${\pm0.4}$                     & 60.4${\pm0.5}$                      & 57.8${\pm0.9}$
                                             & 66.2${\pm0.2}$                    & 58.9${\pm0.3}$                      & 67.1${\pm0.5}$                     & 62.3${\pm0.4}$                      & \colorbox{gray}{OOM}    & \colorbox{gray}{OOM}    & \colorbox{gray}{OOM} & \colorbox{gray}{OOM}                                  \\
            \midrule
            GCN+\textsc{Guard}
                                             & \textbf{81.7${\pm0.3}$}
                                             & 75.5${\pm0.4}$
                                             & 64.8${\pm0.6}$                    & 71.3${\pm0.4}$                      & 56.4${\pm0.2}$
                                             & \textbf{77.6${\pm0.2}$}
                                             & \textbf{72.7${\pm0.4}$}
                                             & \textbf{77.1${\pm0.3}$}           & \textbf{71.9${\pm0.2}$}             & \textbf{62.3${\pm0.3}$}            & 60.9${\pm0.5}$                      & \textbf{89.8${\pm0.4}$} & \textbf{87.9${\pm0.6}$}                                                                                \\
            SGC+\textsc{Guard}
                                             & 81.4${\pm0.2}$
                                             & \textbf{79.9${\pm0.5}$}
                                             & \textbf{72.9${\pm0.6}$}           & \textbf{78.7${\pm0.7}$}             & \textbf{59.4${\pm0.2}$}
                                             & 76.7${\pm0.2}$
                                             & 71.5${\pm0.5}$
                                             & 76.8${\pm0.3}$                    & 70.4${\pm0.3}$                      & 62.0${\pm0.8}$                     & \textbf{61.7${\pm0.4}$}             & 87.2${\pm0.3}$          & 85.6${\pm0.8}$                                                                                         \\
            \bottomrule
        \end{tabular}
    }

\end{table*}

\subsection{Clean Performance}

We will first investigate whether \textsc{Guard} hinders the performance of GCNs in the absence of adversarial attacks.
Table~\ref{tab:clean} summarizes the results on clean datasets. Here we only compare our methods with preprocessing-based defenses for a fair comparison.
Note that SVD requires high computation overhead (typically $\mathcal{O}(N^3)$) for approximating the low-rank components, making it challenging to scale to large datasets. Jaccard is not applicable for arXiv and Reddit since it can only operate on binary node features to calculate similarity scores between nodes. Both approaches applied to GCNs result in a decrease in classification accuracy. In particular, the performance of GCNs has a significant drop when SVD is applied.
One promising property of universal defenses (particularly \textsc{Guard}) we want to remark on is the clean performance. We can observe from Table~\ref{tab:clean} that universal defense methods, when applied on GCNs, have little impact on the clean accuracy of the model. In this regard, we find \textsc{Guard} and \textsc{Deg} to be superior. This is due to the fact that both \textsc{Guard} and \textsc{Deg} tend to identify low-degree nodes as potential attacker nodes, which are usually not connected with the target node in a sparse graph without adversarial attacks. Therefore, the accuracy of GCNs armed with universal defenses is not sacrificed in benign situations.
Also in terms of scalability and flexibility, we find \textsc{Guard} to be superior compared with SVD and Jaccard. It can easily scale to large datasets like arXiv and Reddit without any constraints and additional overheads.

\subsection{Robustness against Adversarial Attacks}
\subsubsection{Robustness compared with preprocessing-based defenses.}
In Table \ref{tab:after_attack}, we present the experimental results against different adversarial attacks.
Note that Nettack is a greedy-based exhaustive attack that is not feasible for datasets larger than Cora. Additionally, both FGA and IG require a dense adjacency matrix to compute the approximated gradients of each edge, which throw an out-of-memory error when applied to larger datasets arXiv and Reddit.
We found that \textsc{Guard} can successfully defend against adversarial attacks for GCNs and achieves the best results in most cases. The results suggest that the adversarial attacks are ``reversible'' by a well-designed defensive patch. Also, one can see that \textsc{Deg} achieved a good performance on Cora and Reddit, reflecting the attacker's tendency to pick low-degree nodes to craft adversarial edges. On PubMed and arXiv, two datasets exhibit a more significant long-tail distribution, \textsc{Deg} did not perform as well as expected, which indicates that simply identifying anchor nodes based on node degrees can not ensure the attacker nodes are included. By contrast, \textsc{Guard} is more stable and achieves the best performance in most cases.
Although \textsc{Guard} is built upon the empirical observations on gradient-based attacks, we can also see that \textsc{Guard} is still effective in defending against Nettack, which is a non-gradient-based attack. This demonstrates the versatility and generalizability of \textsc{Guard} as a defense mechanism.

\subsubsection{Robustness compared with model-based defenses.}
Given the promising results demonstrating the increased robustness of GCNs with \textsc{Guard}, we conduct further comparisons with other state-of-the-art robust GCNs. We use vanilla GCN and SGC as backbone models in our experiments.
In Table~\ref{tab:after_attack2}, we present the experimental results on four datasets under different attacks.
The results have shed light on the vulnerability of current defenses, which suffer seriously from adversarial targeted attacks. The results are sobering – most defenses show no or only marginal
improvement compared to an undefended baseline. This is also in line with the results reported in \cite{DBLP:conf/nips/MujkanovicGGB22}
Among the compared methods, MedianGCN and SoftMedian, which leverage median as an aggregation function during message passing, have demonstrated good performance against adversarial attacks. GNNGUARD has achieved the best performance among the baselines by utilizing the attention mechanism to assign smaller weights for adversarial edges. However, they are not able to scale to large graphs.
It is obvious that although the vanilla GCN and SGC are less robust, with the help of \textsc{Guard}, the robustness can be significantly improved and even outperforms the state-of-the-art methods with large margins. Overall, the results provide empirical evidence of our method's superior performance and scalability compared to state-of-the-art approaches.

\begin{figure}[t]
    \centering
    \subfigure[]{\includegraphics[width=0.38\linewidth]{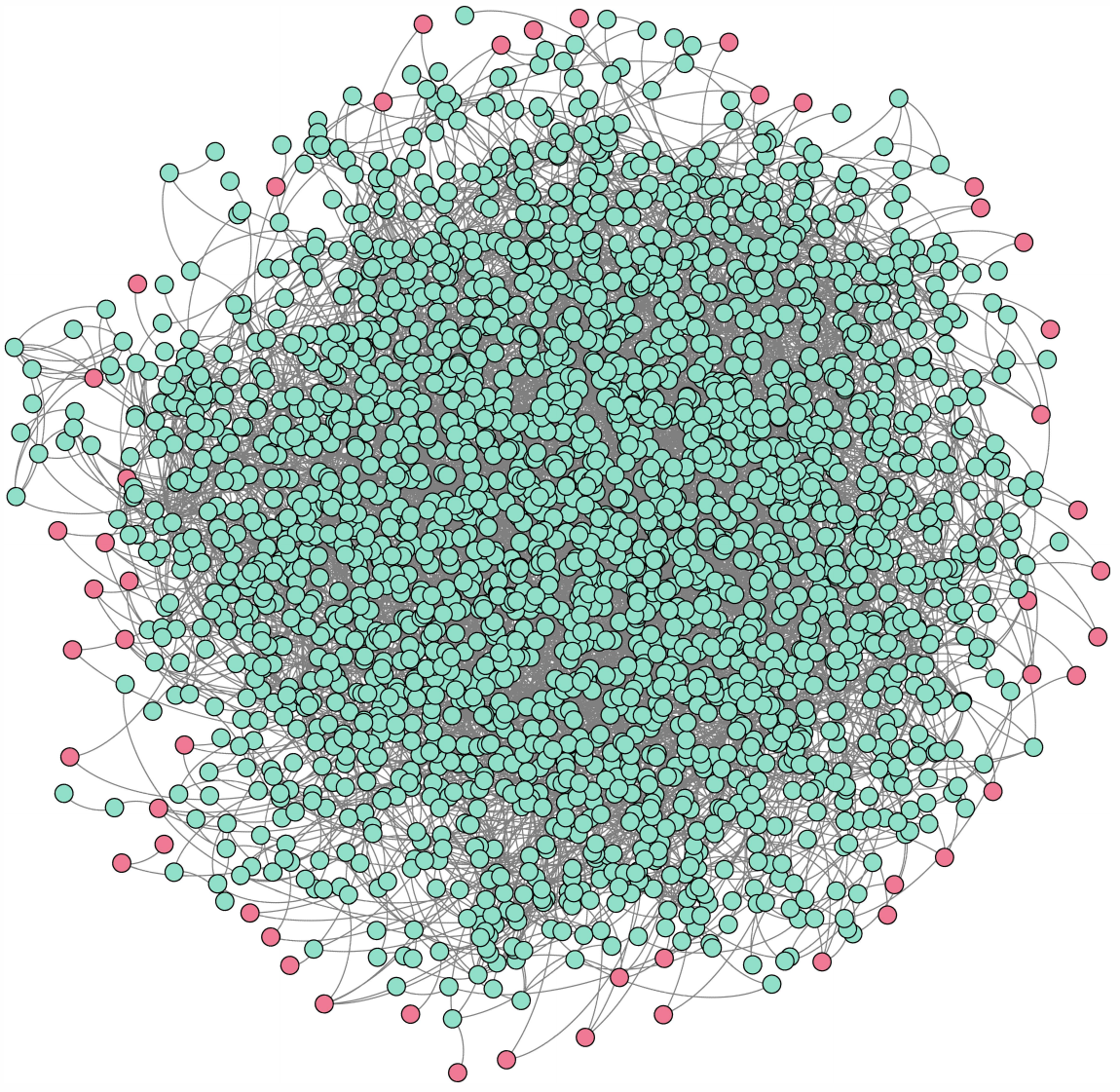}
    \label{fig:vis_a}}
    \subfigure[]{\includegraphics[width=0.45\linewidth]{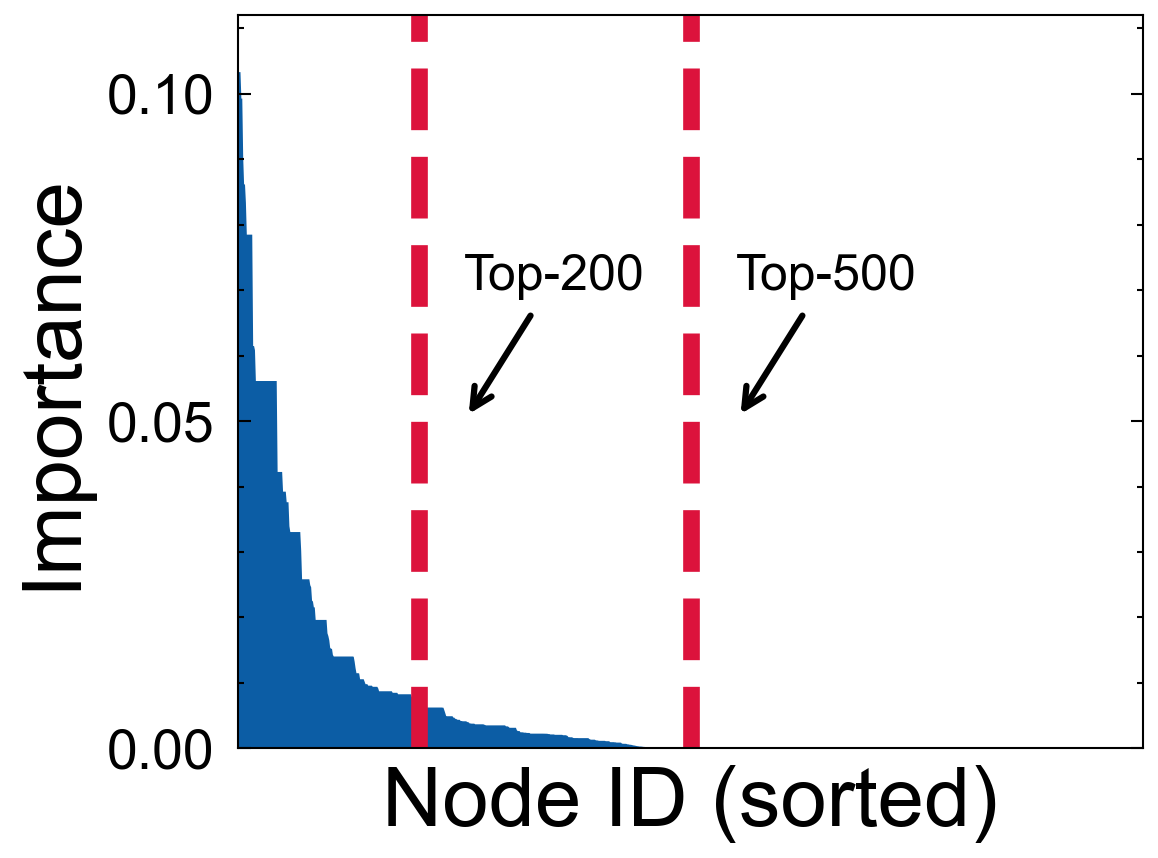}
    \label{fig:vis_b}}
    \caption{\textbf{(a)} Visualization of Cora citation graph with anchor nodes highlighted in {pink}. \textbf{(b)} Influence score of all nodes on Cora dataset. Many anchor nodes have low degrees but contribute significantly to the overall node importance.}
    \label{fig:vis}
\end{figure}

\subsection{Exploratory Results}
\label{sec:explor}

\subsubsection{Visualization on anchor nodes.}
We present the visualization of the Cora citation graph with anchor nodes (here $k=50$) identified by \textsc{Guard} in Figure~\ref{fig:vis_a}. It reveals that the anchor nodes are typically low-degree nodes, which means that most of the nodes in a large and sparse graph will not be connected with anchor nodes. In benign situations, a clean graph is insensitive to \textsc{Guard}, since the connection between a target node and an anchor node which \textsc{Guard} attempts to remove does not exist in most cases. However, malicious attackers tend to exploit these nodes to perturb the target nodes, which makes \textsc{Guard} effective for an attacked graph.

\subsubsection{Influence score.}
According to Proposition~\ref{proposition}, we can also consider the influence score $\mathcal{I}_u^*$ in Eq.~\eqref{eq:influence} as the influence on the GCNs' outputs when connecting node $u$ to any target node. In this regard, a larger $\mathcal{I}_u^*$ would lead to a stronger attack when node $u$ becomes the attacker node. Based on the analysis, we plot the influence score of all nodes on Cora dataset in Figure~\ref{fig:vis_b}, as an illustrative example. The results suggest a long-tailed distribution where the influence score of tailed nodes ranked after 200th is marginal. In other words, GCNs are more sensitive to the top 200 nodes on Cora. Therefore, if we set the anchor set size $k$ as 200 in \textsc{Guard}, most of the attacks on Cora could be defended. This is in line with the results in Table~\ref{tab:after_attack} and further explains why \textsc{Guard} is effective in defending against adversarial targeted attacks.

\begin{table}[t]
    \centering
    \caption{Running time of \textsc{Guard} on four datasets.}\label{tab:time}
    \begin{tabular}{l|cccc}
        \toprule
        \textbf{} & \textbf{Cora} & \textbf{PubMed} & \textbf{arXiv} & \textbf{Reddit} \\
        \midrule
        Size $k$  & 200           & 500             & 10,000         & 20,000          \\
        Time      & 2ms           & 10ms            & 292ms          & 102ms           \\
        \bottomrule
    \end{tabular}
\end{table}

\subsubsection{Efficiency of \textsc{Guard}.}
As a universal defense method, \textsc{Guard} also enjoys high efficiency and low complexity given its superior effectiveness as demonstrated in previous sections.
Table \ref{tab:time} shows the running time of \textsc{Guard} on four datasets. The running time reported here does not include the training time of the surrogate model. We can see that the computation of anchor nodes is efficient once the surrogate model is given. The running time is quite acceptable even on large datasets arXiv and Reddit. Moreover, \textsc{Guard} can be trivially parallelized and accelerate the computations on larger graphs. Note that the anchor nodes are only computed once to form the universal patch, which further shows the superiority of our method in terms of efficiency.

\begin{table}[t]
    \centering
    \caption{Performance of \textsc{Guard} on Cora and Pubmed datasets under various attacks. The results are averaged over five runs.}
    \label{tab:victims}
    \resizebox{\linewidth}{!}{
        \begin{tabular}{l|cccc|cccc}
            \toprule
                    & \multicolumn{4}{c}{\textbf{Cora}} & \multicolumn{4}{c}{\textbf{PubMed}}                                                                                                                                   \\
            \cmidrule{2-9}
                    & \textbf{GAT}                      & \textbf{\textsc{+Guard}}            & \textbf{JKNet} & \textbf{\textsc{+Guard}} & \textbf{GAT} & \textbf{\textsc{+Guard}} & \textbf{JKNet} & \textbf{\textsc{+Guard}} \\
            \midrule
            {Clean} & 83.3                              & \colorbox{gray}{83.2}               & 83.2           & \colorbox{gray}{83.3}    & 84.8         & \colorbox{gray}{84.8}    & 83.9           & \colorbox{gray}{83.7}    \\
            \midrule
            SGA     & 29.2                              & \colorbox{gray}{82.1}               & 6.2            & \colorbox{gray}{81.5}    & 12.2         & \colorbox{gray}{84.0}    & 4.3            & \colorbox{gray}{74.1}    \\
            FGA     & 36.9                              & \colorbox{gray}{79.4}               & 16.2           & \colorbox{gray}{79.2}    & 12.4         & \colorbox{gray}{78.5}    & 8.1            & \colorbox{gray}{75.0}    \\
            IG      & 29.0                              & \colorbox{gray}{75.9}               & 12.3           & \colorbox{gray}{69.2}    & 17.0         & \colorbox{gray}{83.9}    & 28.3           & \colorbox{gray}{83.2}    \\
            RBCD    & 25.4                              & \colorbox{gray}{73.5}               & 9.3            & \colorbox{gray}{76.7}    & 10.4         & \colorbox{gray}{82.1}    & 6.2            & \colorbox{gray}{77.5}    \\
            \bottomrule
        \end{tabular}
    }
\end{table}

\subsubsection{Performance on other GCNs.}
To make our results more convincing, we conduct additional experiments on GAT and JKNet.
The results are summarized in Table~\ref{tab:victims}, which reveal similar observations: (i) \textsc{Guard} effectively enhances the robustness of GAT and JKNet without compromising their clean performance. (ii) \textsc{Guard} significantly outperforms undefended baselines against a variety of attacks. These observations further demonstrate the generalizability and effectiveness of \textsc{Guard} in improving the robustness of various GCNs against adversarial attacks. Most importantly, the proposed universal defense can be a valuable addition to existing defenses, providing a reliable defense against adversarial attacks in real-world applications.

\subsubsection{Hyper-parameter analysis.}
\begin{figure}[t]
    \centering
    \includegraphics[width=0.48\linewidth]{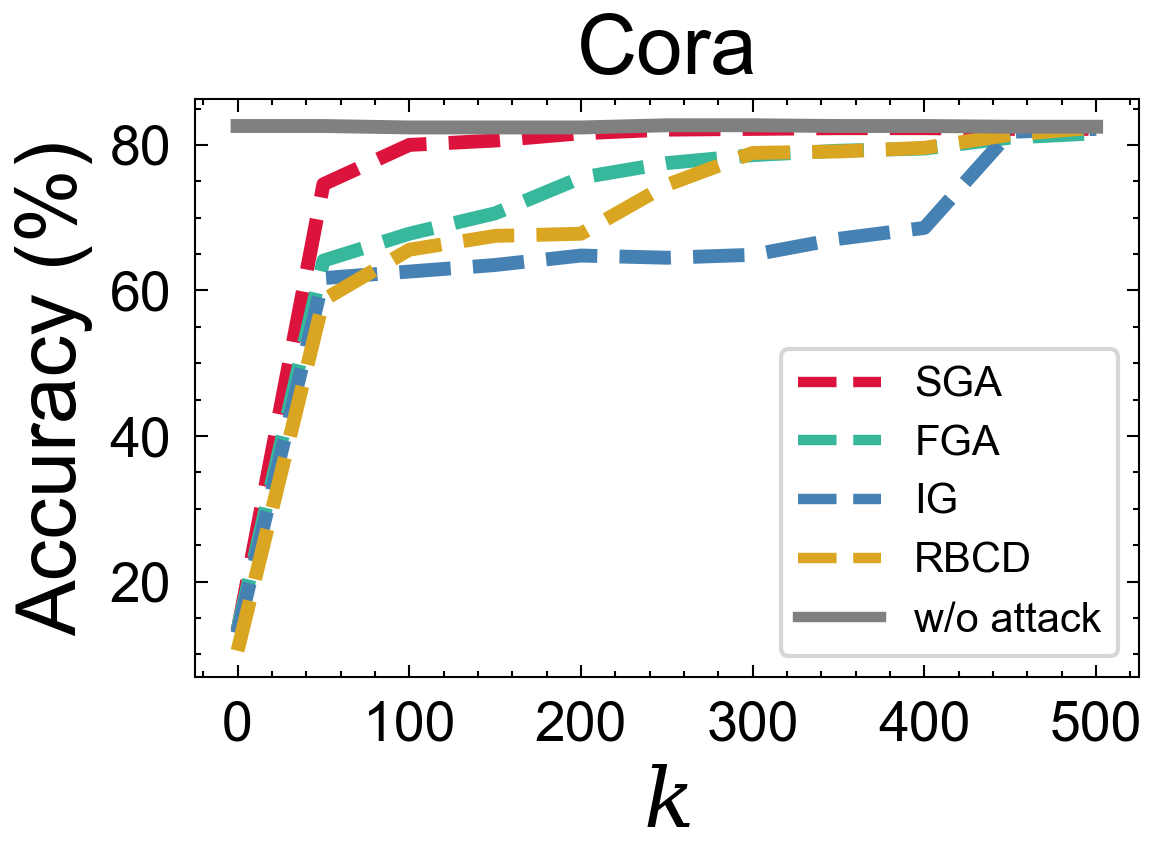}
    \includegraphics[width=0.48\linewidth]{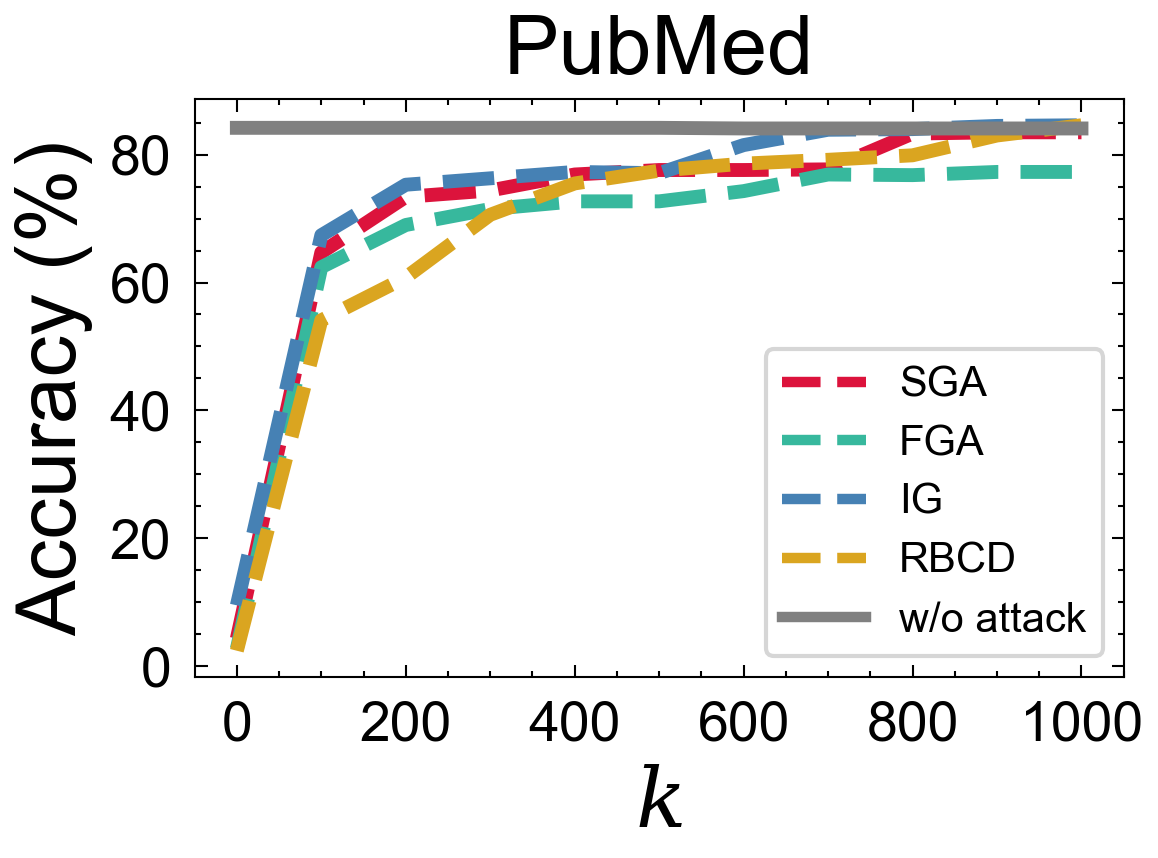}
    \includegraphics[width=0.48\linewidth]{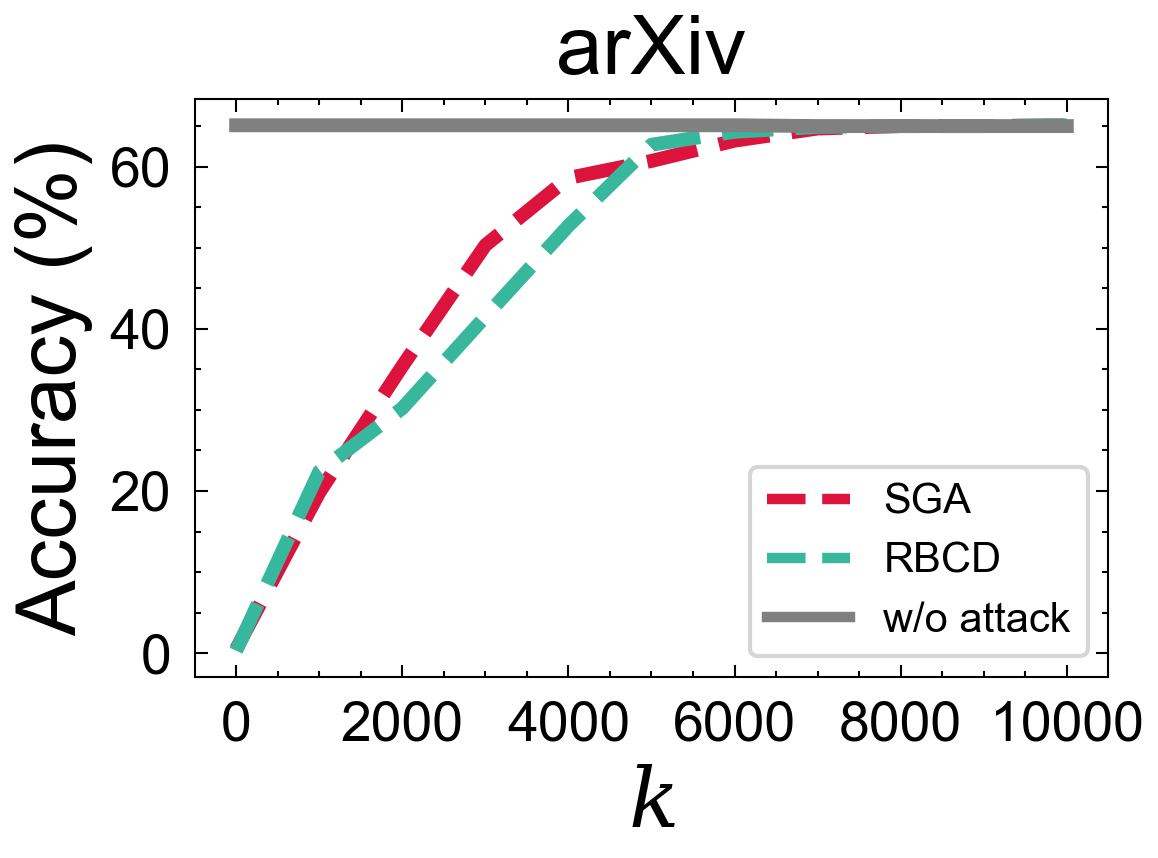}
    \includegraphics[width=0.48\linewidth]{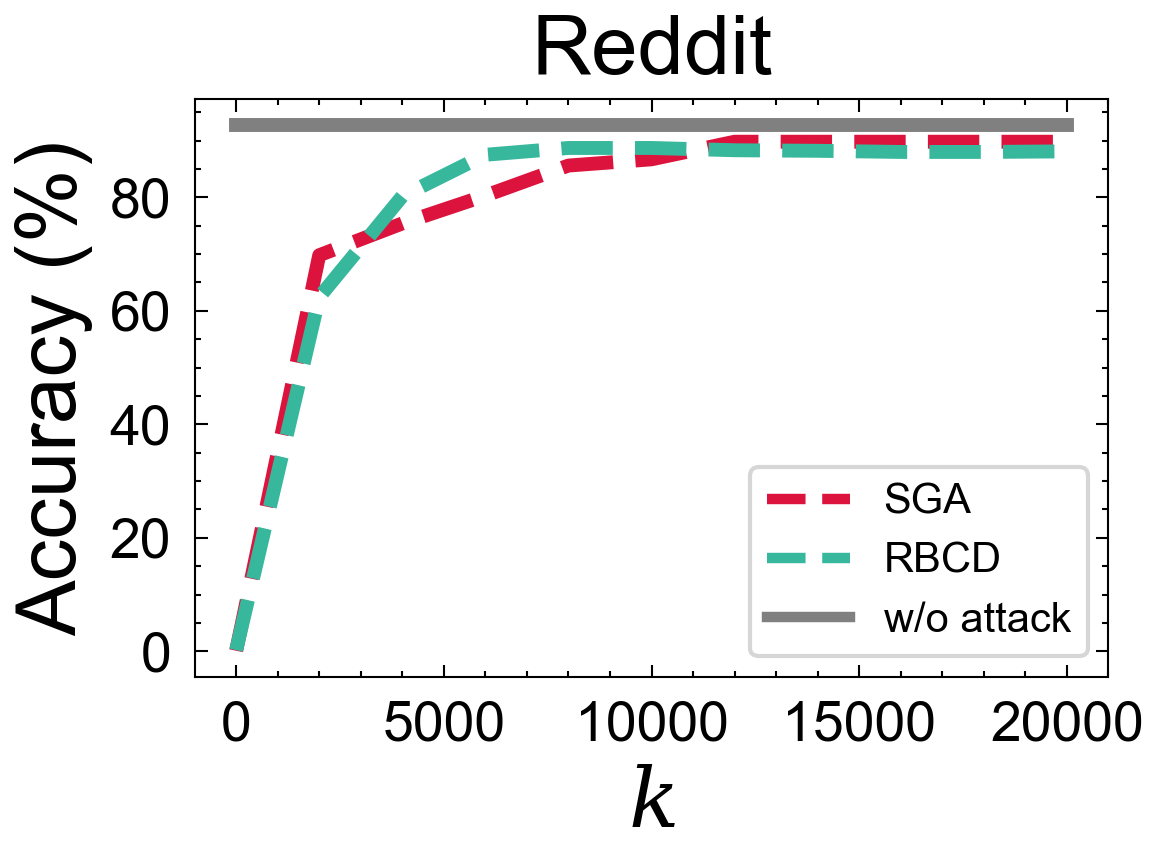}
    \caption{The accuracy of GCN+\textsc{Guard} on four datasets with varying the number of anchor nodes $k$, repeated five times.}
    \label{fig:abla_k}
\end{figure}

\begin{figure}[t]
    \centering
    \includegraphics[width=0.48\linewidth]{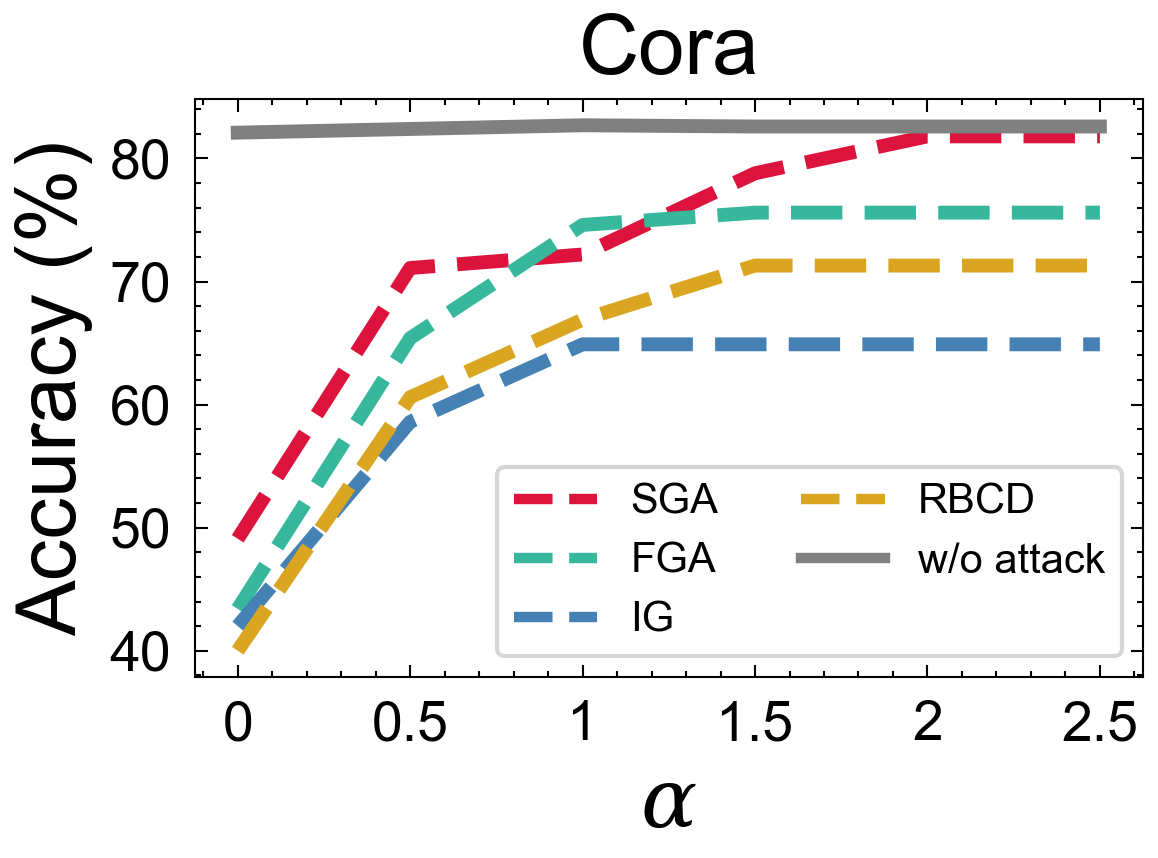}
    \includegraphics[width=0.48\linewidth]{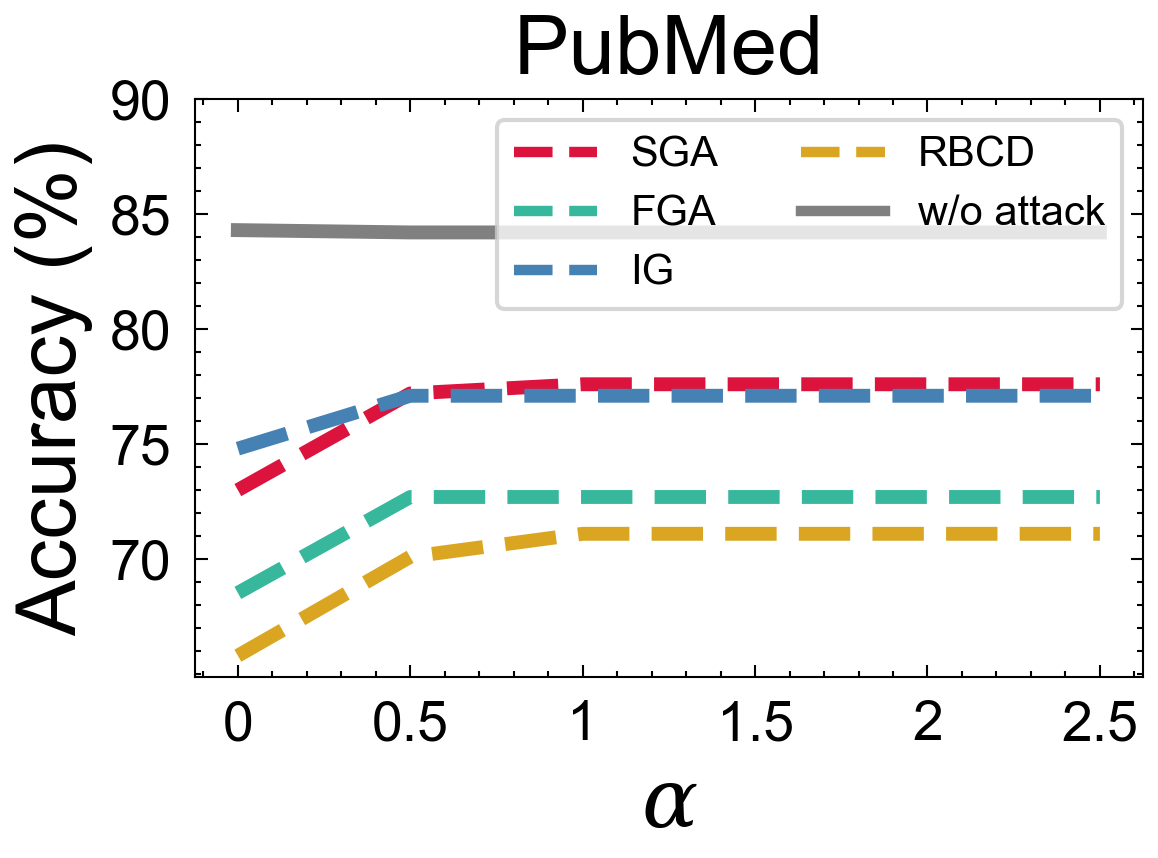}
    \includegraphics[width=0.48\linewidth]{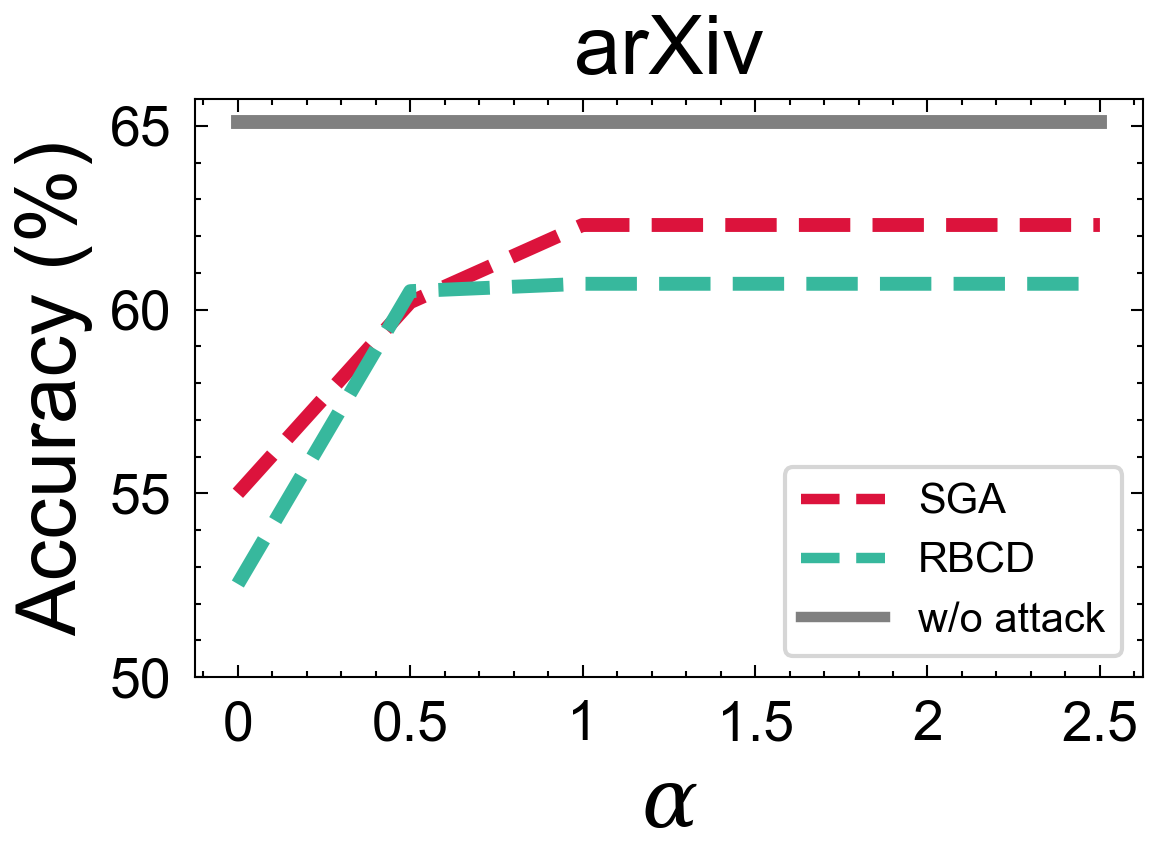}
    \includegraphics[width=0.48\linewidth]{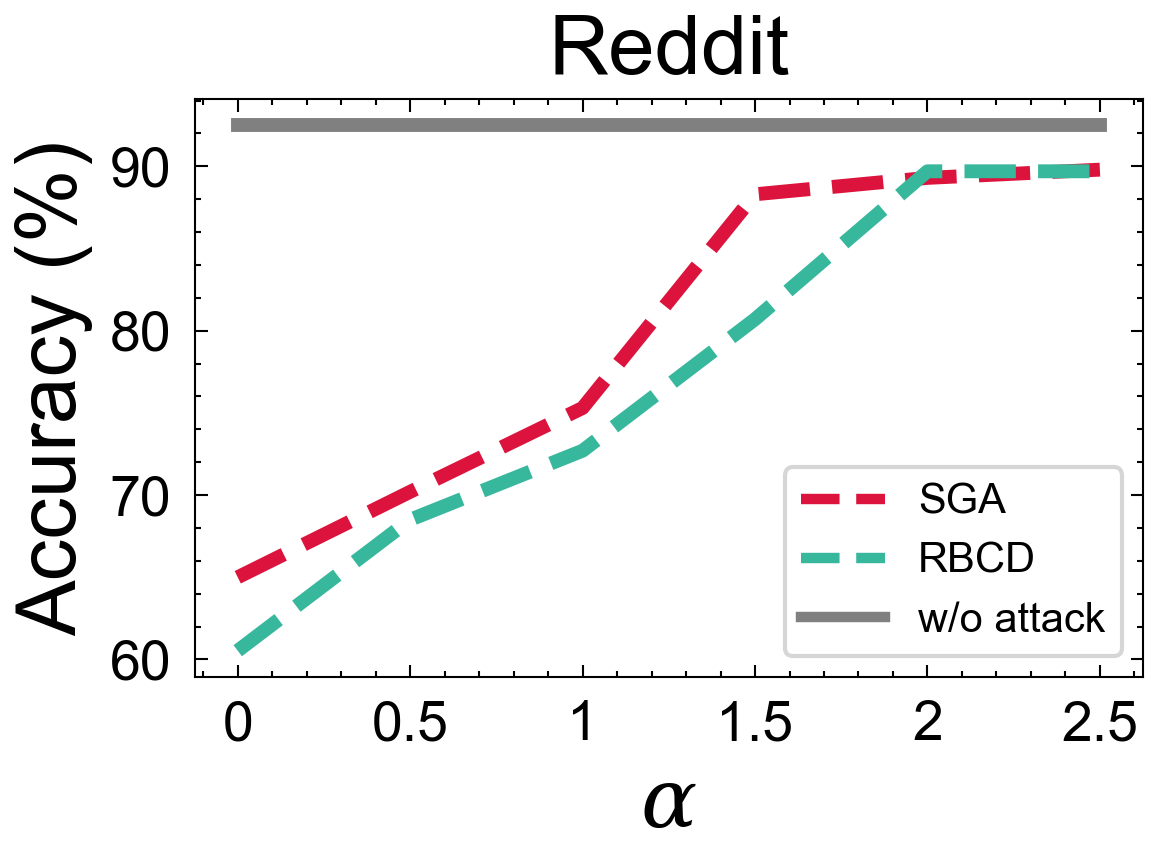}
    \caption{The accuracy of GCN+\textsc{Guard} on four datasets with varying the scaling factor $\alpha$, repeated five times.}
    \label{fig:abla_alpha}
\end{figure}

We perform case studies on four datasets to qualitatively evaluate the impact of $k$ and $\alpha$ in \textsc{Guard}, respectively. We report the performance of GCN with \textsc{Guard} by varying one and fixing another as the optimal value. As shown in Figure \ref{fig:abla_k}, a larger value of $k$ leads to a larger set of anchor nodes and thus improves the robustness of GCN. When $k$ reaches a critical value, \textit{e.g.}, 450 on Cora, the accuracy of GCN against attacks reaches the clean accuracy, which means the adversarial attacks are successfully defended. In most cases, the performance of GCN changes smoothly on the clean graph, which indicates that GCN is not sensitive to \textsc{Guard} in benign situations.
Observed from Figure \ref{fig:abla_alpha}, we can see that \textsc{Guard} is more sensitive to $\alpha$ on Cora and Reddit, as evidenced by increasing $\alpha$ can significantly improve the performance of \textsc{Guard} and the best performance is achieved when $\alpha=2$.
In contrast, \textsc{Guard} is less sensitive to $\alpha$ on PubMed and arXiv.
The difference may be due to the different sparsity of the two datasets.
Overall, the results are consistent with our findings in Figure~\ref{fig:degree} as the malicious edges are typically those connected with low-degree nodes. These findings highlight the importance of identifying low-degree nodes as potential targets for adversarial attacks in advance.

\section{Conclusion and Future work}
\label{sec:conclusion}
In this work, for the first time, we study a novel problem of protecting GCNs from adversarial targeted attacks with a \textit{universal} defense method. We demonstrate that attackers tend to perturb target nodes with a fixed set of low-degree nodes. We analyze possible reasons to explain the finding. Our understanding motivates us to propose \textsc{Guard}, a simple yet effective method to enhance the robustness of GCNs. Specifically, \textsc{Guard} generates a universal patch by explicitly identifying the possible attacker nodes and pruning all suspicious edges in advance to protect important local nodes from multiple adversarial targeted attacks. In our extensive experimental evaluation, \textsc{Guard} can successfully defend against various strong attacks, achieving state-of-the-art robustness without sacrificing accuracy when applied to several established GCNs.

Despite the promising experimental justifications, our method might potentially suffer from some limitations: (i) Currently, we only consider the purification in the context of structural perturbations, however, adversarial perturbations will occur at the level of features and nodes. The defense strategy should be adapted to meet the requirements to defend against other forms of adversarial attacks (\textit{e.g.,} attacks on node features).
(ii) Our work is specific to the node classification task, although it is also possible to extend the proposed method in the future to other node-related tasks (\textit{e.g.,} link prediction) by suitably modifying the loss function. We leave these for future work.

\begin{acks}
The research is supported by the National Key R\&D Program of China under grant No. 2022YFF0902500, the Guangdong Basic and Applied Basic Research Foundation, China (No. 2023A1515011050), Ant Group through Ant Research Program (20210002).
\end{acks}

\bibliographystyle{ACM-Reference-Format}
\bibliography{main}

\end{document}